\documentclass[lettersize,journal]{IEEEtran}
\usepackage{amsmath,amsfonts}
\usepackage{algorithmic}
\usepackage{algorithm}
\usepackage{array}
\usepackage[caption=false,font=normalsize,labelfont=sf,textfont=sf]{subfig}
\usepackage{textcomp}
\usepackage{stfloats}
\usepackage{url}
\usepackage{verbatim}
\usepackage{graphicx}
\usepackage{cite}
\usepackage{multirow}
\usepackage{makecell}
\usepackage{CJKutf8}
\usepackage{pifont}
\usepackage{threeparttable}
\usepackage{graphicx}

\usepackage{color}
\usepackage[table]{xcolor}
\usepackage{colortbl}
\usepackage{diagbox}

\begin{document}

\title{ViLReF: An Expert Knowledge Enabled \\ Vision-Language Retinal Foundation Model}

\author{Shengzhu Yang, Jiawei Du, Jia Guo, Weihang Zhang, \\ Hanruo Liu, Huiqi Li,~\IEEEmembership{Senior Member,~IEEE}, and Ningli Wang
\thanks{

This research work is supported by the National Natural Science Foundation
of China (NSFC) (Grant No. 82572313 and Grant No. 62506036) and  Hebei Natural Science Foundation (Grant No. F2025105040).

Shengzhu Yang, Jiawei Du, Weihang Zhang, and Huiqi Li are with Beijing Institute of Technology and Beijing Key Laboratory of Intelligent Diagnosis Technology and Equipment for Optic Nerve-Related Eye Diseases.

Jia Guo is with Tsinghua University.

Hanruo Liu is with Beijing Tongren Hospital, Capital Medical University; Beijing Key Laboratory of Intelligent Diagnosis Technology and Equipment for Optic Nerve-Related Eye Diseases; Henan Eye Hospital, Henan Provincial People's Hospital.

Ningli Wang is with Henan Academy of Innovations in Medical Science; Beijing Key Laboratory of Intelligent Diagnosis Technology and Equipment for Optic Nerve-Related Eye Diseases; Beijing Tongren Hospital, Capital Medical University.

Corresponding authors: Weihang Zhang (zhangweihang@bit.edu.cn), Huiqi Li (huiqili@bit.edu.cn), and Ningli Wang (wningli@vip.163.com).

}}

\markboth{IEEE TRANSACTIONS ON IMAGE PROCESSING,~Vol.~xx, No.~xx, September~2025}%
{Shell \MakeLowercase{\textit{et al.}}: A Sample Article Using IEEEtran.cls for IEEE Journals}


\maketitle

\begin{abstract}
    Subtle semantic differences in retinal image and text data present great challenges for pre-training visual-language models. Moreover, false negative samples, i.e., image-text pairs having the same semantics but incorrectly regarded as negatives, disrupt the visual-language pre-training process and affect the model's learning ability. This work aims to develop a retinal foundation model, called ViLReF, by pre-training on a paired dataset comprising 451,956 retinal images and corresponding diagnostic text reports. In our vision-language pre-training strategy, we leverage expert knowledge to facilitate the extraction of labels and propose a novel constraint, the Weighted Similarity Coupling Loss, to adjust the speed of pushing sample pairs further apart dynamically within the feature space. Furthermore, we employ a batch expansion module with dynamic memory queues, maintained by momentum encoders, to supply extra samples and compensate for the vacancies caused by eliminating false negatives. Extensive experiments are conducted on multiple datasets for downstream classification and segmentation tasks. The experimental results demonstrate the powerful zero-shot and transfer learning capabilities of ViLReF, verifying the effectiveness of our pre-training strategy. Our ViLReF model is available at: \url{https://github.com/T6Yang/ViLReF}.
\end{abstract}


\section{Introduction}
\label{intro}

\IEEEPARstart{F}{oundation} models pre-trained on large datasets have shown impressive generalization capabilities when fine-tuned on various downstream tasks \cite{moor2023foundation, radford2021learning}. With the availability of training data of ophthalmology, retinal foundation models are now widely applied in clinical practice. Retinal images and their corresponding diagnostic reports have been employed for training foundation models. Unlike natural images, retinal images often differ only in subtle pathological regions  \cite{wang2022medclip}, making contrastive self-supervised learning particularly challenging.

\begin{figure}[!t]
    \centerline{\includegraphics[width=\columnwidth]{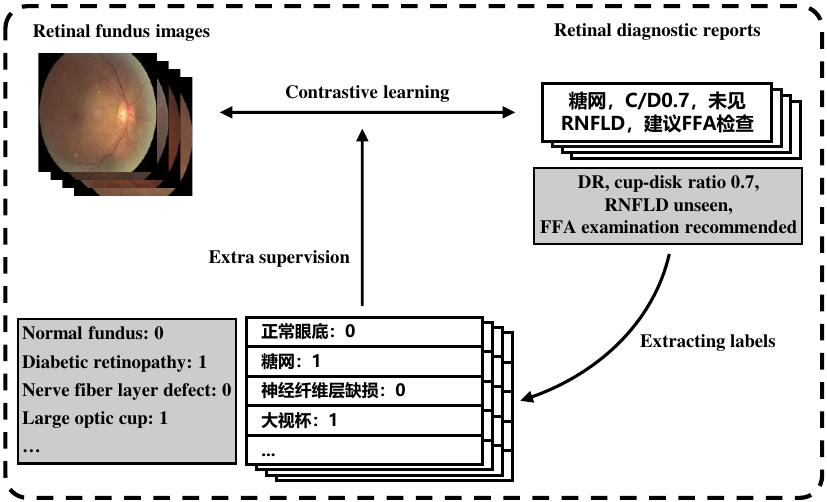}}
    \caption{Overview of the proposed vision-language pre-training strategy. Labels are extracted from Chinese retinal diagnostic reports to provide extra supervision for image-text contrastive learning, mitigating the effect caused by false negative samples. Note: The text in the gray box is the English translation of the adjacent Chinese text.}
    \label{fig1}
\end{figure}

Self-supervised contrastive vision-language joint representation learning is a common strategy to train foundation models, which typically does not require additional manual annotations except for the diagnostic reports \cite{huang2021gloria}. 
However, traditional self-supervised contrastive learning strategies may treat image-text pairs having the same or similar semantics (e.g., one case indicates mild diabetic retinopathy while another case indicates diabetic retinopathy) as negative samples (i.e., false negative samples) and push them apart during pre-training, which introduces noise, obscures the appearance of pathological differences to models, and leads to confusion in learning clinically relevant visual representations. One way to mitigate the effect caused by false negative samples is to introduce extra supervision during pre-training.

\IEEEpubidadjcol

According to the text reports, a series of labels can be extracted automatically to provide supervision and avoid the occurrence of false negatives. The Unified Medical Language System (UMLS) \cite{bodenreider2004unified} recognizes clinically defined entities within English text reports to extract labels conveniently 
However, plenty of institutions use retinal diagnostic reports written in Chinese, and there is no Chinese UMLS available, and existing methods for identifying Chinese entities using UMLS predominantly rely on translation and mapping. The quality of translation affects the results of mapping significantly, thereby limiting the efficiency and effectiveness of mapping algorithms \cite{chen2023mapping}. Moreover, some entities cannot be converted into labels after being identified directly, thus they require further logical decisions based on expert knowledge. Therefore, to accommodate large and diverse retinal diagnostic reports in Chinese and ensure accurate and effective label generation, a new method to acquire supervision labels from Chinese retinal text reports is needed.

To this end, we present an expert-knowledge-based report converter for data pre-processing to convert Chinese text reports into supervision labels, and propose a novel pre-training strategy for retinal foundation models, as illustrated in Fig. \ref{fig1}. A novel constraint in the feature space is proposed, which leverages labels to adjust the speed of pushing sample pairs apart dynamically. This mitigates the effect of false negative samples during pre-training. However, it also poses a new challenge: the number of contrastive samples might be reduced, affecting the pre-training process.

To compensate for the weakened contrastive learning effect resulting from the elimination of false negatives, memory queues are maintained by momentum encoders to expand the quantity of contrastive samples. Ultimately, we propose ViLReF, an expert knowledge enabled \textbf{Vi}sion-\textbf{L}anguage \textbf{Re}tinal \textbf{F}oundation model, which comprises two aligned universal retinal image and text feature extractors. The main contributions of this work can be summarized as follows:
\begin{itemize}
    \item{We propose ViLReF, a foundation model pre-trained on a dataset of 451,956 paired retinal images and corresponding diagnostic reports provided by Beijing Tongren Hospital. Utilizing contrastive learning without reliance on pretext tasks during pre-training, ViLReF effectively understands and aligns rich visual and language semantic representations of ophthalmic data. Our vision-language pre-training strategy helps the model capture subtle but clinically significant visual patterns in the data.}
    \item{To address the challenge of false negative samples while ensuring effective representation learning during pre-training, we introduce a novel contrastive learning pre-training strategy. Specifically, we develop the expert-knowledge-based report converter, which leverages domain knowledge to extract labels from Chinese retinal diagnostic text reports. Additionally, we propose the Weighted Similarity Coupling Loss to optimize the coupling distribution of feature and label similarities, aiding the model in learning robust representations. Furthermore, we employ a batch expansion module with dynamic memory queues to compensate for the absent contrastive samples caused by the elimination of false negatives.}
    \item{We extensively evaluate ViLReF's inference performance on various downstream tasks across multiple datasets. Compared with existing pre-training strategies, our method demonstrates impressive effectiveness, endowing ViLReF with excellent semantic understanding and generalization capabilities in the ophthalmic domain.}
\end{itemize}

\section{Related Work}
\subsection{Vision-language Pre-trained Foundation Models}

Pre-training enables foundation models to learn transferable data representations, improving robustness and generalization capabilities compared with training from scratch. Current pre-training strategies can be generally divided into two categories: learning from auxiliary pretext tasks and contrastive learning.

Auxiliary pretext tasks are typically achieved through self-supervision, generating pseudo labels to help the model extract representative knowledge \cite{jaiswal2020survey}. For example, InsLoc \cite{yang2021instance} builds a detection pretext task by clipping and synthesizing images. Super-resolution frameworks such as HIPA \cite{cai2023hipa} and NasmamSR \cite{yang2022nasmamsr} are also employed for representation extraction. In retinal domain, foundation models like RETFound \cite{zhou2023foundation} and EyeFound \cite{shi2024eyefound} adopt masked autoencoders (MAE) \cite{he2022masked} learning information-dense image representations by predicting masked regions, demonstrating state-of-the-art (SOTA) performance in disease prediction, diagnosis, and prognosis.

Contrastive learning offers a straightforward method for learning representations by pulling features of positive samples closer and pushing features of negative samples further apart. Sample pairs are selected during pre-training dynamically \cite{ericsson2022self}. For example, SimCLR \cite{chen2020simple} promotes alignment between augmented views of an image through a non-linear projection and contrastive loss. To reduce InfoNCE's sensitivity to batch size, MoCo \cite{he2020momentum} uses momentum-updated encoder outputs as the embedding repository. DINO \cite{caron2021emerging} enhances the representative consistency across rescaled image crops using self-distillation. SwAV \cite{caron2020unsupervised} and PCL \cite{li2020prototypical} cluster the data to avoid computationally intensive pairwise comparisons and reduce false negatives.

Contrastive pre-training strategies, exemplified by CLIP \cite{radford2021learning}, which combine vision and language to match positive and negative sample pairs automatically, are growing in popularity. Among medical data sources, images and their associated text descriptions are two common and abundant data modalities. The emergence of large-scale medical datasets such as MIMIC-CXR \cite{johnson2019mimic}, PadChest \cite{bustos2020padchest} and ROCO \cite{pelka2018radiology}, provides necessary data to pre-train medical foundation models. 

Recently, various studies have explored contrastive pre-trained vision-language foundation models in medical domain. ConVirt \cite{zhang2022contrastive} processes image patches and sentence snippets as semantically aligned views. BioMedCLIP \cite{zhang2023large} utilizes 15 million image-text pairs for pre-training, achieving strong performance on various downstream tasks. GLoRIA \cite{huang2021gloria} applies global and sub-region losses to extract multi-scale representations. In retinal domain, RetiZero\cite{wang2024common} leverages 341,896 image-text pairs sourced from public datasets, ophthalmic literature, and online resources, covering over 400 retinal diseases, while FLAIR \cite{silva2025foundation} and KeepFIT \cite{wu2024mm} are pre-trained on datasets constructed from 37 open-access sources with synthesized textual descriptions. KeepFIT further leverages a synthetic dataset comprising 1 million images covering 14 diseases to enhance pre-training. To reflect real-world clinical scenarios, RET-CLIP \cite{du2024ret} utilizes binocular image-text datasets to learn monocular and patient-level representations. EyeCLIP \cite{shi2025multimodal} adopts a hierarchical keyword extraction strategy that standardizes clinical report expressions and organizes them into structured text input, leveraging textual information for contrastive learning effectively. However, it primarily captures surface-level terms and does not incorporate deeper clinical reasoning or explicit label-based supervision.

\subsection{Expert Knowledge as Additional Supervision}

Compared with the natural domain, medical data exhibits finer-grained, denser, and more specialized semantics. In ophthalmological diagnosis through retinal images, interpretative descriptions characterize structural appearances (e.g., optic disk color, vessel orientation, nerve fiber thickness) and local lesions (e.g., exudation, retinopathy, neovascularization). Diagnostic reports serve as the clinical decision, indicating conditions such as arteriosclerosis, diabetic retinopathy (DR), and glaucoma. While interpretive and diagnostic contents are presented at different levels, they are highly interdependent and predominantly associated with expert knowledge. 

Medical tasks driven by expert knowledge typically follow agent-based paradigms. For instance, observations of exudation, hemorrhage, neovascularization, and retinal damage can indicate DR \cite{sivaprasad2023perspectives}. Drusen size and pigment epithelium changes are correlated to age-related retinopathy staging \cite{fleckenstein2024age}. Expert knowledge also serves as clinical priors: exudate segmentation indicates macular edema severity \cite{giancardo2012exudate}, and the optic cup-disk ratio supports glaucoma diagnosis \cite{lu2023pkrt}.

Contrastive learning foundation models pre-trained on image and text data, such as CLIP, often fail to fully capitalize on the expert knowledge contained in medical data, risking the loss of fine-grained representations during the pre-training process. Additionally, false negative sample pairs can reduce the model performance significantly. To address these challenges, existing strategies have leveraged expert knowledge to obtain additional supervision. For example, MedCLIP \cite{wang2022medclip} uses rule-based label extraction from clinical texts and decouples image-text associations, expanding pre-training data and reducing false negatives. MedKLIP \cite{wu2023medklip} enhances representation learning by incorporating clinical expert knowledge and disease descriptions. In retinal domain, FLAIR \cite{silva2025foundation} maps labels into text descriptions using expert knowledge, addressing the scarcity of text-based supervision in public retinal datasets. KeepFIT\cite{wu2024mm} leverages high-quality image-text pairs collected from professional fundus diagram books. It employs image similarity-guided text revision and a mixed pre-training strategy to infuse expert knowledge.

In contrastive representation learning within retinal domain, we leverage expert knowledge to extract diagnostic labels from detailed clinical medical data as additional supervision, thereby mitigating the effect of false negatives and enabling the model to understand finer-grained pathological structures and learn accurate clinical reasoning.

\begin{figure*}[!t]
    \centering
    \centerline{\includegraphics[width=0.811\textwidth]{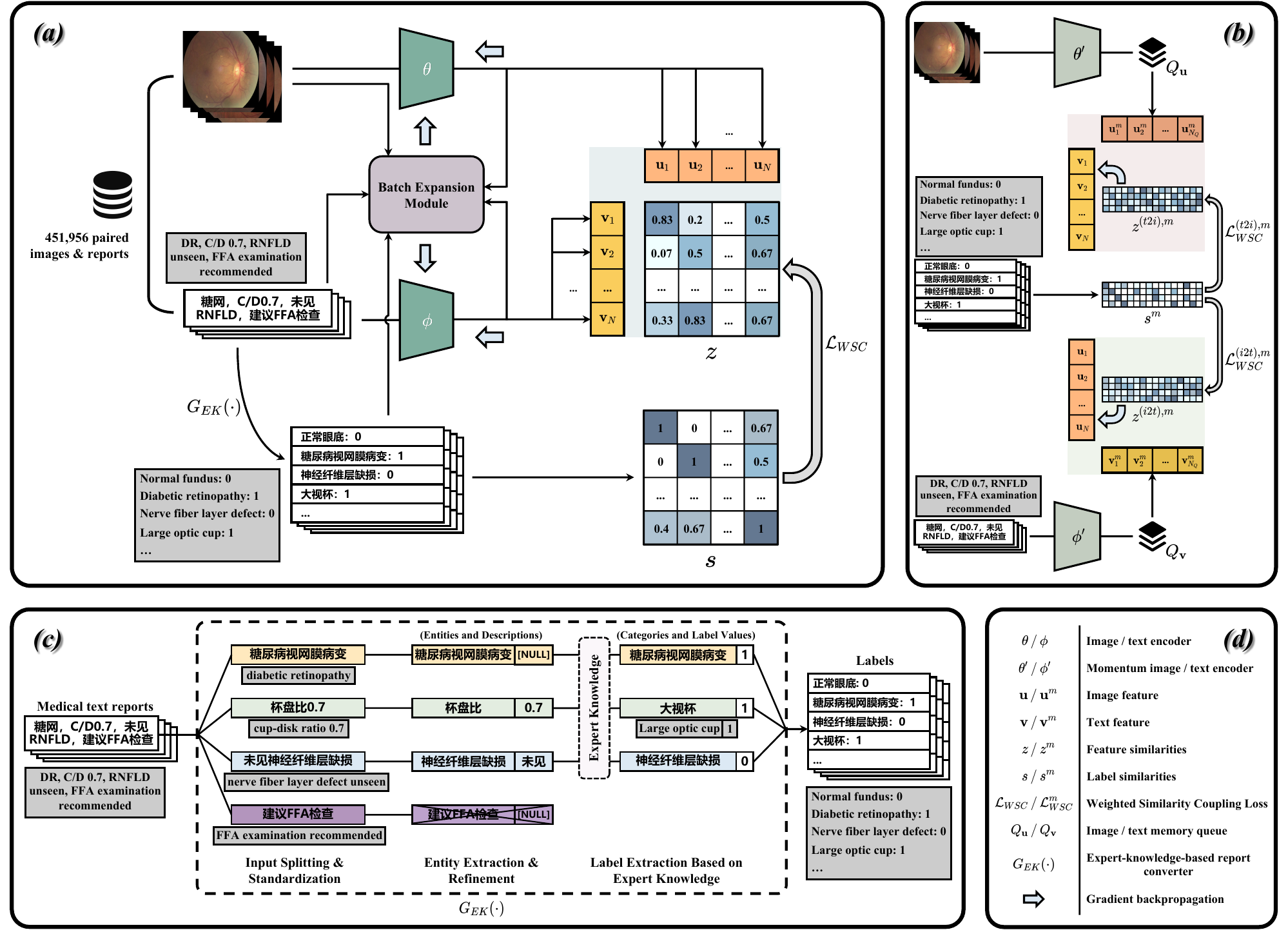}}
    \caption{Flowchart of our vision-language pre-training strategy. (a) The backbone of the ViLReF model pre-training. (b) Illustration of the batch expansion module. (c) Illustration of the expert-knowledge-based report converter. (d) Legend of symbols presented in the flowchart. Note: The text in the gray box is the English translation of the adjacent Chinese text.}
    \label{fig2}
\end{figure*}

\section{Method}
In this section, we introduce the pre-training process and technical details of our model. The workflow is illustrated in Fig. \ref{fig2}(a). After sampling a batch of retinal image-text pairs from the dataset, we first extract labels from the text report using the proposed expert-knowledge-based report converter $G_{EK}(\cdot)$. Features are then extracted by the image encoder $\theta(\cdot)$ and the text encoder $\phi(\cdot)$. We use the proposed Weighted Similarity Coupling Loss $\mathcal{L}_{WSC}$ to constrain the feature similarities of images and texts based on the prior similarities between their labels. Meanwhile, we employ a batch expansion module with dynamic memory queues, maintained by the momentum encoders $\theta'(\cdot)$ and $\phi'(\cdot)$, to effectively expand the equivalent batch size.

\subsection{Label Extraction Driven by Expert Knowledge}

The extraction of labels for additional supervision, guided by rules built on expert knowledge, is a pre-processing step before pre-training. We propose an expert-knowledge-based report converter $G_{EK}(\cdot): \mathcal{T}^{l \times 1} \mapsto \mathcal{Y}^{c \times 1}$, as shown in Fig. \ref{fig2}(c), to convert texts into labels, where $T_n \in \mathcal{T}^{l \times 1}$ represents a Chinese diagnostic report with a maximal length $l$, and $y_n \in \mathcal{Y}^{c \times 1}$ is a binary label sequence containing $c$ categories obtained by $G_{EK}(T_n)$. $G_{EK}(\cdot)$ consists of three modules: Input Splitting \& Standardization, Entity Extraction \& Refinement, and Label Extraction Based on Expert Knowledge. 

\begin{CJK}{UTF8}{gbsn}
When a retinal diagnostic report is input into $G_{EK}(\cdot)$, it is first split into multiple phrases, and the Chinese vocabulary is standardized to facilitate entity mapping, ensuring accurate label extraction. During this step, common abbreviations used by clinicians are standardized into full medical terms. For example, ``糖网" (DR) is standardized to ``糖尿病视网膜病变" (Diabetic retinopathy), and ``RNFLD" is standardized to ``神经纤维层缺损" (Nerve fiber layer defect). Then, to refine the semantic information, entities with their descriptions are extracted, and irrelevant phrases (e.g., medical advice such as ``FFA (fluorescein fundus angiography) examination recommended") that are not related to the image information are filtered out. Finally, clinical expert knowledge is utilized to make decisions based on entities and their descriptions to obtain the labels of phrases. For example, ``cup-disk ratio more than 0.5" can be inferred as ``large optic cup", and ``arteriovenous ratio smaller than 2:3" can be inferred as ``thin arteries". The combination of labels from multiple phrases constitutes the label set of each retinal diagnostic report. Since the retinal images and diagnostic reports are in pairs, the obtained text labels can also be regarded as the annotations of the corresponding retinal images.
\end{CJK}

To set the number of required labels, we use the method described in Luo et al. \cite{luo2019pkuseg} to split the standardized texts and count word frequencies in the entire text domain. Finally, we select the ``normal” category, several ``abnormal" categories, and the ``others” category. The ``abnormal" categories include both disease categories (such as ``diabetic retinopathy”, ``fundus arteriosclerosis” and ``cataract”) and lesion categories (such as ``cotton wool spots”, ``exudation” and ``macular pucker”). Diseases and lesions with very low word frequencies are grouped into the category of ``others" because the probability of encountering false negative samples within a mini-batch of contrastive learning is extremely low.

\subsection{Feature Extraction and Similarity Calculation}

Given a large dataset $\mathcal{D}$, we sample a subset $\mathcal{D}_s = \{ ( I_n, T_n, y_n ) \}_{n=1}^N \subset \mathcal{D}$ as the mini-batch input, where $I_n \in \mathbb{R}^{H \times W \times C}$ represents a retinal image. Our model consists of an image encoder $\theta(\cdot)$ and a text encoder $\phi(\cdot)$, each containing a feature extractor and a nonlinear projection layer. The projection layer aims to map the extracted features from a narrow band to the entire space, effectively avoiding the feature collapse phenomenon \cite{gupta2022understanding}.

The features $\mathbf{u}_n = \theta(I_n)$ and $\mathbf{v}_n = \phi(\hat{T}_n)$ are obtained after inputting the image and tokenized report sequence $\hat{T}_n$ to the image and text encoders. Then, the cosine similarity $z_{i,j}$ between image and text features can be calculated as: 
\begin{equation}
    \label{eq. 1}
    z_{i,j} = \frac {\mathbf{u}_i \cdot \mathbf{v}_j^\top} {{\| \mathbf{u}_i \|}_2 \cdot {\| \mathbf{v}_j \|}_2}.
\end{equation}

Similarly, according to the obtained labels, the label similarity $s_{i,j}$ can be calculated as: 
\begin{equation}
    \label{eq. 2}
    s_{i,j} =
    \begin{cases}
        \hfil \displaystyle\frac{y_i \cdot y_j^\top}{{\| y_i \|}_2 \cdot {\| y_j \|}_2} & {\| y_i \|}_2 \cdot {\| y_j \|}_2 \neq 0, \vspace{2ex}\\
        \vspace{1ex}
        \hfil 0                                                                      & {\| y_i \|}_2 \cdot {\| y_j \|}_2 = 0.
    \end{cases}
\end{equation}
Note that when calculating the cosine similarity of labels, the ``others" category is not included, which explains the situation where ${\| y_i \|}_2 \cdot {\| y_j \|}_2 = 0$. In our model, the samples of ``others" class are negative samples relative to every other sample in the same pre-training mini-batch.

\subsection{Speed Adjustment of Feature Similarity}
\label{SA}

Inspired by InfoNCE Loss, we propose the Weighted Similarity Coupling Loss $\mathcal{L}_{WSC}$, which aims to supply a novel constraint, adjusting the speed of pushing features further apart within a mini-batch. This loss function can utilize the additional supervision provided by labels appropriately, making the pre-training process more robust. For the ``image to text (i2t)" and ``text to image (t2i)'' conditions, the proposed $\mathcal{L}_{WSC}^{(i2t)}$ and $\mathcal{L}_{WSC}^{(t2i)}$ are formulated as:

\begin{equation}
    \label{eq. 3}
    \mathcal{L}_{WSC}^{(i2t)} = - \frac{1}{N} \sum_{i}^{N} \log \left[\frac{\sigma_{i,i}}{\sigma_{i,i} + \sum_{j \neq i}^{N} \left( 1 - s_{i,j} \right) \sigma_{i,j}} \right],  
\end{equation}

\begin{equation}
    \label{eq. 4}
    \mathcal{L}_{WSC}^{(t2i)} = - \frac{1}{N} \sum_{i}^{N} \log \left[\frac{\sigma_{i,i}}{\sigma_{i,i} + \sum_{j \neq i}^{N} \left( 1 - s_{i,j} \right) \sigma_{j,i}} \right], 
\end{equation}
where $\sigma_{i,i} = \exp (z_{i,i} / \tau )$ and $\sigma_{i,j} = \exp (z_{i,j} / \tau )$ are the exponential self-similarity and mutual-similarity between image and text features in the mini-batch, and $\tau$ denotes the temperature coefficient. By coupling $\sigma_{i,i}$ and $\sigma_{i,j}$ using $s_{i,j}$ to establish their correlation, we increase the similarities between positive samples while reducing them between negatives.

Specifically, $s_{i,j} = 0$ indicates that the label similarity between a positive sample and a true negative sample in a mini-batch is completely different. In this case, the weight for their feature similarity $(1 - s_{i,j})$ is $1$, which is equivalent to performing a softmax operation and pushing the true negative sample from the positive sample as far as possible. When $s_{i,j} = 1$, it indicates that the two samples have completely identical labels, and the weight for their feature similarity $(1 - s_{i,j})$ is 0, causing the false negative sample to be ignored in the calculation. When $0 < s_{i,j} < 1$, the feature similarity $\sigma _{i,j}$ is linearly weighted by the label similarity between the two samples, adjusting the speed of pushing their features further apart dynamically. 

It can be observed that $\mathcal{L}_{WSC}$ pulls the positive samples closer in the feature space, while the speed at which the features of other samples are pushed further apart is adjusted by $s_{i,j}$. Specifically, taking $\mathcal{L}_{WSC}^{(i2t)}$ as an example, its negative gradient with respect to $\sigma_{i,i}$ is given by the partial derivative: 

\begin{equation}
    \label{eq. 5}
    - \nabla_{\sigma_{i,i}} \mathcal{L}_{WSC}^{(i2t)} = \frac{1}{N} \cdot \frac{ \sum_{j \neq i}^{N} { \left( 1 - s_{i,j} \right) \sigma_{i,j}}}{ \sigma_{i,i} \cdot \left[ \sigma_{i,i} + \sum_{j \neq i}^{N} { \left( 1 - s_{i,j} \right) \sigma_{i,j}} \right] },
\end{equation}
where the calculation result is always non-negative, indicating that the positive samples are always pulled closer in the feature space. Similarly, the partial derivative of $\mathcal{L}_{WSC}^{(i2t)}$ with respect to $\sigma_{i,j \neq i}$ is given by:

\begin{equation}
    \label{eq. 6}
    - \nabla_{\sigma_{i,j \neq i}} \mathcal{L}_{WSC}^{(i2t)} = - \frac{1}{N} \cdot \frac{1 - s_{i,j}}{ \sigma_{i,i} + \sum_{j \neq i}^{N} { \left( 1 - s_{i,j} \right) \sigma_{i,j}}}, 
\end{equation}
which showcases that the absolute gradient decreases with the increase of $s_{i,j}$. When $s_{i,j}$ is high, indicating similar samples, the gradient becomes small, thus pushing similar samples further apart more slowly in the feature space. In contrast, the gradient is larger when $s_{i,j}$ is low, pushing samples apart faster.

\subsection{Batch Expansion Using Memory Queues}
Eliminating or linearly weighting the contrastive samples can influence $\mathcal{L}_{WSC}$, thereby reducing the gradients and weakening the pre-training effect. To address this issue, we employ a batch expansion module using dynamic memory queues to compensate for the absent contrastive samples, as illustrated in Fig. \ref{fig2}(b).

First, the image and text encoders are augmented with extra momentum encoders $\theta '(\cdot)$ and $\phi '(\cdot)$, respectively. The parameters of momentum encoders are initialized as their corresponding image and text encoders, and they are not updated via gradients during pre-training. Instead, they are updated using the momentum updating approach:
\begin{equation}
    \label{eq. 7}
    \begin{aligned}
        \displaystyle p(\theta ')_t \, {\rm{ \xleftarrow{} }} & \, m \cdot p(\theta ')_{t-1} + (1 - m) \cdot p(\theta)_{t}, \\
        \displaystyle p(\phi ')_t \, {\rm{ \xleftarrow{} }}   & \, m \cdot p(\phi ')_{t-1} + (1 - m) \cdot p(\phi)_{t},     \\
        p(\theta ')_0 \; {\rm{ = }}                           & \; p(\theta)_0,                                           \\
        p(\phi ')_0 \; {\rm{ = }}                             & \; p(\phi)_0,
    \end{aligned}
\end{equation}
where $p(\cdot)$ denotes the network parameters and $m$ is the preset momentum coefficient. The variable $t$ represents the current global pre-training step. 

The momentum encoders extract and store features into two dynamic memory queues, $Q_{\mathbf{u}}$ and $Q_{\mathbf{v}}$, each with the size of $N_{Q}$. Then, we calculate the exponential feature similarities $\sigma_{i,j}^{(i2t),m} = \exp (z_{i,j}^{(i2t),m} / \tau )$ and $\sigma_{j,i}^{(t2i),m} = \exp (z_{j,i}^{(t2i),m} / \tau )$ between the features $\mathbf{u}_{i}$ ($\mathbf{v}_{i}$) and the features $\mathbf{v}_j^{m}$ ($\mathbf{u}_j^{m}$) in $Q_{\mathbf{v}}$ ($Q_{\mathbf{u}}$), where

\begin{equation}
    \label{eq. 8}
    z_{i,j}^{(i2t),m} = \frac {\mathbf{u}_i \cdot \mathbf{v}_j^{m \top}} {{\| \mathbf{u}_i \|}_2 \cdot {\| \mathbf{v}_j^m \|}_2},
\end{equation}

\begin{equation}
    \label{eq. 9}
    z_{j,i}^{(t2i),m} = \frac {\mathbf{u}_j^m \cdot \mathbf{v}_i^\top} {{\| \mathbf{u}_j^m \|}_2 \cdot {\| \mathbf{v}_i \|}_2}. 
\end{equation}
Similar to Eq. (\ref{eq. 2}), we calculate the label similarity as:

\begin{equation}
    \label{eq. 10}
    s_{i,j}^m =
    \begin{cases}
        \hfil \displaystyle\frac{y_i \cdot y_j^{m \top}}{{\| y_i \|}_2 \cdot {\| y_j^m \|}_2} & {\| y_i \|}_2 \cdot {\| y_j^m \|}_2 \neq 0, \vspace{2ex}\\
        \vspace{1ex}
        \hfil 0                                                                      & {\| y_i \|}_2 \cdot {\| y_j^m \|}_2 = 0.
    \end{cases}
\end{equation}

In this case, the momentum Weighted Similarity Coupling Loss $\mathcal{L}^{(i2t),m}_{WSC}$ and $\mathcal{L}^{(t2i),m}_{WSC}$ are calculated as: 
\begin{multline}
    \label{eq. 11}
    \mathcal{L}^{(i2t),m}_{WSC} = \\ - \frac{1}{N} \sum_{i}^{N} \log \left[\frac{\sigma_{i,i}^{(i2t),m}}{\sigma_{i,i}^{(i2t),m} + \sum_{j \neq i}^{N_{Q}} \left( 1 - s_{i,j}^{m} \right) \sigma_{i,j}^{(i2t),m}} \right], 
\end{multline}

\begin{multline}
    \label{eq. 12}
    \mathcal{L}^{(t2i),m}_{WSC} = \\ - \frac{1}{N} \sum_{i}^{N} \log \left[\frac{\sigma_{i,i}^{(t2i),m}}{\sigma_{i,i}^{(t2i),m} + \sum_{j \neq i}^{N_{Q}} \left( 1 - s_{i,j}^{m} \right) \sigma_{j,i}^{(t2i),m}} \right]. 
\end{multline}

This module expands the equivalent batch size by using $Q_{\mathbf{u}}$ and $Q_{\mathbf{v}}$ effectively, compensating for the weakened pre-training effect caused by the gradient reduction of loss functions, while the computational overhead is only slightly increased by momentum encoders. The slow update of momentum encoders also ensures sequential relevance of each feature in $Q_{\mathbf{u}}$ and $Q_{\mathbf{v}}$, avoiding inconsistencies in feature discrimination caused by rapid changes.

Finally, the overall loss function can be formulated as: 
\begin{equation}
    \label{eq. 13}
    \mathcal{L} = \mathcal{L}^{(i2t)}_{WSC} + \mathcal{L}^{(t2i)}_{WSC} + \mathcal{L}^{(i2t), m}_{WSC} + \mathcal{L}^{(t2i), m}_{WSC}.
\end{equation}

\subsection{Implementation Settings}
\label{impledetails}
We use PyTorch to build the network and pre-train ViLReF based on Chinese CLIP \cite{yang2022chinese} (CN-CLIP). We utilize ViT-B/16 \cite{dosovitskiy2020image} as the image encoder and RoBERTa-wwm-ext-base-chinese \cite{cui2020revisiting} as the text encoder. Retinal images are resized to $224 \times 224$, and flipped horizontally with probability $0.5$. Since positional descriptions in ophthalmic diagnostic reports are typically based on anatomical quadrants relative to the nasal and temporal sides of the retina (rather than absolute left–right orientation), horizontal flipping does not alter the semantic consistency between retinal images and their corresponding reports. A color jitter factor $0.1$ is used to adjust brightness, contrast, and saturation. The Chinese text input is segmented by characters rather than words \cite{li2019word}, 
with maximum length $l = 100$. The batch size $N$ is set to $256$, the projection layer outputs are fixed at $512$ dimensions, and the temperature coefficient $\tau$ is initialized to $0.07$, following the original CLIP \cite{radford2021learning} setting. The momentum coefficient $m$ and memory queue size $N_Q$ are set to $0.75$ and $768$ via grid search based on the lowest loss on the validation set. AdamW \cite{loshchilov2017decoupled} is used with learning rate $3e-5$, $\beta_1 = 0.9$, $\beta_2 = 0.98$, $\epsilon = 1e-6$, and weight decay $\lambda = 0.001$. 

The early stopping criterion is defined based on a validation set of 1,000 class-balanced and potentially multi-label samples drawn from the training data (without any overlap with the training set). The epoch with the lowest validation loss on this subset is selected as the stopping point. The pre-training time on a single RTX 3090 GPU is 16 hours using automatic mixed precision training. In evaluating downstream tasks for comparing the image representation learning capability, to leverage richer information, we use the features extracted by the feature extractors rather than the projection layers. 

\section{Experiments}
\label{sec:guidelines}
In this section, we validate our model on various downstream tasks and datasets to demonstrate the superior performance of ViLReF and the effectiveness of the pre-training strategy. We detail the pre-training and evaluation datasets, present experimental results comparing our pre-training strategy with existing ones, conduct an ablation study to verify the contribution of each component, and compare the performance of our model with SOTA retinal foundation models.

\subsection{Datasets}

\subsubsection{Pre-training dataset}
The pre-training dataset comprises 451,956 pairs of retinal images and corresponding diagnostic reports provided by Beijing Tongren Hospital. The retinal images were collected from various medical institutions across China, and the diagnostic reports were authored by professional ophthalmologists. Patients' private information was removed during data pre-processing. We analyzed the frequency of Chinese words in the text data, removed semantic ambiguities, and identified 33 common categories, including 1 ``normal" category, 7 diseased categories, 24 lesion categories, and 1 ``others" category containing extremely rare categories. The category labels are represented in a multi-hot binary format. This approach is based on the hypothesis that the likelihood of samples with rare categories appearing in the same mini-batch is very low,  which makes them reasonable to be simply classified into an “others” category and will not influence the pre-training as false negative samples. 

\subsubsection{Evaluation datasets}
For evaluation, we use eight public datasets: RFMiD \cite{pachade2021retinal}, ODIR \cite{challenge2019peking}, REFUGE \cite{orlando2020refuge}, MESSIDOR \cite{decenciere2014feedback}, FIVES \cite{jin2022fives}, IDRiD \cite{porwal2018indian}, Retinal-Lesions \cite{icpr2020-LesionNet}, and FGADR \cite{9257400}. 
This selection of datasets provides a comprehensive evaluation of our model across a wide range of downstream tasks with different retinal conditions.

\subsection{Evaluation Metrics}
For quantitative study, we adopt the Area Under the receiver operating characteristic Curve (AUC) and mean Average Precision (mAP) as the classification evaluation metrics. AUC evaluates overall performance, while mAP focuses more on evaluating long-tailed label data. For evaluating segmentation results, we use the Dice Similarity Coefficient (DSC) and the Intersection over Union (IoU). DSC measures the similarity between the segmentation result and the ground truth, while IoU calculates the overlap between the segmentation result and the ground truth. The reported values for each experiment are the mean and standard deviation over five repeated runs. Throughout the experiments, we assess statistical significance using paired \textit{t}-tests. All improvements reported are statistically significant with p $<$ 0.05.

\begin{table*}[!t]
    \begin{center}
    \caption{Evaluation on Linear Probing Performance for Different Pre-training Strategies. Each Encoder is Fixed to ViT-B/16. (\%)}
    \renewcommand{\arraystretch}{1.1}
    \label{table2}
    \begin{tabular}{l c c c c c c c c c c c c}
    \hline \hline
    \multirow{2}*{\makecell[l]{Pre-training \\ Strategy}} & \multicolumn{2}{c}{RFMiD} & \multicolumn{2}{c}{ODIR} &
    \multicolumn{2}{c}{REFUGE} & \multicolumn{2}{c}{MESSIDOR} & \multicolumn{2}{c}{FIVES} & \multicolumn{2}{c}{\textit{Avg.}}\\
    \cline{2-13}
    \multirow{2}*{} & AUC & mAP & AUC & mAP & AUC & mAP & AUC & mAP & AUC & mAP & AUC & mAP \\
    \hline
    \multirow{2}*{\makecell[l]{MAE}} & \multirow{2}*{\makecell[l]{79.57\\ \; \scalebox{0.9}[0.9]{(0.03)}}} & \multirow{2}*{\makecell[l]{24.59\\ \; \scalebox{0.9}[0.9]{(0.01)}}} & \multirow{2}*{\makecell[l]{77.32\\ \; \scalebox{0.9}[0.9]{(0.01)}}} & \multirow{2}*{\makecell[l]{39.43\\ \; \scalebox{0.9}[0.9]{(0.01)}}} & \multirow{2}*{\makecell[l]{51.40\\ \; \scalebox{0.9}[0.9]{(0.46)}}} & \multirow{2}*{\makecell[l]{50.11\\ \; \scalebox{0.9}[0.9]{(0.16)}}} & \multirow{2}*{\makecell[l]{56.49\\ \; \scalebox{0.9}[0.9]{(0.45)}}} & \multirow{2}*{\makecell[l]{31.03\\ \; \scalebox{0.9}[0.9]{(0.26)}}} & \multirow{2}*{\makecell[l]{74.36\\ \; \scalebox{0.9}[0.9]{(0.19)}}} & \multirow{2}*{\makecell[l]{51.57\\ \; \scalebox{0.9}[0.9]{(0.15)}}} & \multirow{2}*{\makecell[l]{\textit{67.83}\\ \; \scalebox{0.9}[0.9]{\textit{(0.23)}}}} & \multirow{2}*{\makecell[l]{\textit{39.35}\\ \; \scalebox{0.9}[0.9]{\textit{(0.12)}}}} \\
    \multirow{2}*{\makecell[l]{}} & \multirow{2}*{\makecell[l]{}} & \multirow{2}*{\makecell[l]{}} & \multirow{2}*{\makecell[l]{}} & \multirow{2}*{\makecell[l]{}} & \multirow{2}*{\makecell[l]{}} & \multirow{2}*{\makecell[l]{}} & \multirow{2}*{\makecell[l]{}} & \multirow{2}*{\makecell[l]{}} & \multirow{2}*{\makecell[l]{}} & \multirow{2}*{\makecell[l]{}} & \multirow{2}*{\makecell[l]{}} & \multirow{2}*{\makecell[l]{}} \\
    \multirow{2}*{\makecell[l]{CLIP}} & \multirow{2}*{\makecell[l]{92.63\\ \; \scalebox{0.9}[0.9]{(0.11)}}} & \multirow{2}*{\makecell[l]{60.34\\ \; \scalebox{0.9}[0.9]{(0.13)}}} & \multirow{2}*{\makecell[l]{91.86\\ \; \scalebox{0.9}[0.9]{(0.15)}}} & \multirow{2}*{\makecell[l]{68.85\\ \; \scalebox{0.9}[0.9]{(0.32)}}} & \multirow{2}*{\makecell[l]{96.53\\ \; \scalebox{0.9}[0.9]{(0.37)}}} & \multirow{2}*{\makecell[l]{94.41\\ \; \scalebox{0.9}[0.9]{(0.32)}}} & \multirow{2}*{\makecell[l]{82.50\\ \; \scalebox{0.9}[0.9]{(0.38)}}} & \multirow{2}*{\makecell[l]{59.28\\ \; \scalebox{0.9}[0.9]{(0.38)}}} & \multirow{2}*{\makecell[l]{96.33\\ \; \scalebox{0.9}[0.9]{(0.25)}}} & \multirow{2}*{\makecell[l]{90.21\\ \; \scalebox{0.9}[0.9]{(1.00)}}} & \multirow{2}*{\makecell[l]{\textit{91.97}\\ \; \scalebox{0.9}[0.9]{\textit{(0.25)}}}} & \multirow{2}*{\makecell[l]{\textit{74.62}\\ \; \scalebox{0.9}[0.9]{\textit{(0.43)}}}} \\
    \multirow{2}*{\makecell[l]{}} & \multirow{2}*{\makecell[l]{}} & \multirow{2}*{\makecell[l]{}} & \multirow{2}*{\makecell[l]{}} & \multirow{2}*{\makecell[l]{}} & \multirow{2}*{\makecell[l]{}} & \multirow{2}*{\makecell[l]{}} & \multirow{2}*{\makecell[l]{}} & \multirow{2}*{\makecell[l]{}} & \multirow{2}*{\makecell[l]{}} & \multirow{2}*{\makecell[l]{}} & \multirow{2}*{\makecell[l]{}} & \multirow{2}*{\makecell[l]{}} \\
    \multirow{2}*{\makecell[l]{DeiT}} & \multirow{2}*{\makecell[l]{93.03\\ \; \scalebox{0.9}[0.9]{(0.01)}}} & \multirow{2}*{\makecell[l]{57.08\\ \; \scalebox{0.9}[0.9]{(0.09)}}} & \multirow{2}*{\makecell[l]{92.14\\ \; \scalebox{0.9}[0.9]{(0.15)}}} & \multirow{2}*{\makecell[l]{68.33\\ \; \scalebox{0.9}[0.9]{(0.55)}}} & \multirow{2}*{\makecell[l]{96.44\\ \; \scalebox{0.9}[0.9]{(0.25)}}} & \multirow{2}*{\makecell[l]{93.10\\ \; \scalebox{0.9}[0.9]{(0.40)}}} & \multirow{2}*{\makecell[l]{83.21\\ \; \scalebox{0.9}[0.9]{(0.14)}}} & \multirow{2}*{\makecell[l]{59.67\\ \; \scalebox{0.9}[0.9]{(0.31)}}} & \multirow{2}*{\makecell[l]{95.25\\ \; \scalebox{0.9}[0.9]{(0.27)}}} & \multirow{2}*{\makecell[l]{87.74\\ \; \scalebox{0.9}[0.9]{(0.37)}}} & \multirow{2}*{\makecell[l]{\textit{92.01}\\ \; \scalebox{0.9}[0.9]{\textit{(0.16)}}}} & \multirow{2}*{\makecell[l]{\textit{73.18}\\ \; \scalebox{0.9}[0.9]{\textit{(0.34)}}}} \\
    \multirow{2}*{\makecell[l]{}} & \multirow{2}*{\makecell[l]{}} & \multirow{2}*{\makecell[l]{}} & \multirow{2}*{\makecell[l]{}} & \multirow{2}*{\makecell[l]{}} & \multirow{2}*{\makecell[l]{}} & \multirow{2}*{\makecell[l]{}} & \multirow{2}*{\makecell[l]{}} & \multirow{2}*{\makecell[l]{}} & \multirow{2}*{\makecell[l]{}} & \multirow{2}*{\makecell[l]{}} & \multirow{2}*{\makecell[l]{}} & \multirow{2}*{\makecell[l]{}} \\
    \multirow{2}*{\makecell[l]{MedCLIP}} & \multirow{2}*{\makecell[l]{87.16\\ \; \scalebox{0.9}[0.9]{(0.23)}}} & \multirow{2}*{\makecell[l]{42.36\\ \; \scalebox{0.9}[0.9]{(0.28)}}} & \multirow{2}*{\makecell[l]{91.36\\ \; \scalebox{0.9}[0.9]{(0.01)}}} & \multirow{2}*{\makecell[l]{66.13\\ \; \scalebox{0.9}[0.9]{(0.19)}}} & \multirow{2}*{\makecell[l]{96.07\\ \; \scalebox{0.9}[0.9]{(0.29)}}} & \multirow{2}*{\makecell[l]{92.71\\ \; \scalebox{0.9}[0.9]{(0.20)}}} & \multirow{2}*{\makecell[l]{81.85\\ \; \scalebox{0.9}[0.9]{(0.20)}}} & \multirow{2}*{\makecell[l]{59.32\\ \; \scalebox{0.9}[0.9]{(0.39)}}} & \multirow{2}*{\makecell[l]{92.03\\ \; \scalebox{0.9}[0.9]{(0.24)}}} & \multirow{2}*{\makecell[l]{83.31\\ \; \scalebox{0.9}[0.9]{(0.29)}}} & \multirow{2}*{\makecell[l]{\textit{89.69}\\ \; \scalebox{0.9}[0.9]{\textit{(0.14)}}}} & \multirow{2}*{\makecell[l]{\textit{68.77}\\ \; \scalebox{0.9}[0.9]{\textit{(0.27)}}}} \\
    \multirow{2}*{\makecell[l]{}} & \multirow{2}*{\makecell[l]{}} & \multirow{2}*{\makecell[l]{}} & \multirow{2}*{\makecell[l]{}} & \multirow{2}*{\makecell[l]{}} & \multirow{2}*{\makecell[l]{}} & \multirow{2}*{\makecell[l]{}} & \multirow{2}*{\makecell[l]{}} & \multirow{2}*{\makecell[l]{}} & \multirow{2}*{\makecell[l]{}} & \multirow{2}*{\makecell[l]{}} & \multirow{2}*{\makecell[l]{}} & \multirow{2}*{\makecell[l]{}} \\
    \multirow{2}*{\makecell[l]{UniCL}} & \multirow{2}*{\makecell[l]{91.07\\ \; \scalebox{0.9}[0.9]{(0.24)}}} & \multirow{2}*{\makecell[l]{54.41\\ \; \scalebox{0.9}[0.9]{(0.16)}}} & \multirow{2}*{\makecell[l]{90.28\\ \; \scalebox{0.9}[0.9]{(0.11)}}} & \multirow{2}*{\makecell[l]{65.12\\ \; \scalebox{0.9}[0.9]{(0.15)}}} & \multirow{2}*{\makecell[l]{94.17\\ \; \scalebox{0.9}[0.9]{(0.57)}}} & \multirow{2}*{\makecell[l]{92.64\\ \; \scalebox{0.9}[0.9]{(0.46)}}} & \multirow{2}*{\makecell[l]{80.14\\ \; \scalebox{0.9}[0.9]{(0.21)}}} & \multirow{2}*{\makecell[l]{59.33\\ \; \scalebox{0.9}[0.9]{(0.32)}}} & \multirow{2}*{\makecell[l]{93.29\\ \; \scalebox{0.9}[0.9]{(0.12)}}} & \multirow{2}*{\makecell[l]{82.76\\ \; \scalebox{0.9}[0.9]{(0.23)}}} & \multirow{2}*{\makecell[l]{\textit{89.79}\\ \; \scalebox{0.9}[0.9]{\textit{(0.25)}}}} & \multirow{2}*{\makecell[l]{\textit{70.85}\\ \; \scalebox{0.9}[0.9]{\textit{(0.26)}}}} \\
    \multirow{2}*{\makecell[l]{}} & \multirow{2}*{\makecell[l]{}} & \multirow{2}*{\makecell[l]{}} & \multirow{2}*{\makecell[l]{}} & \multirow{2}*{\makecell[l]{}} & \multirow{2}*{\makecell[l]{}} & \multirow{2}*{\makecell[l]{}} & \multirow{2}*{\makecell[l]{}} & \multirow{2}*{\makecell[l]{}} & \multirow{2}*{\makecell[l]{}} & \multirow{2}*{\makecell[l]{}} & \multirow{2}*{\makecell[l]{}} & \multirow{2}*{\makecell[l]{}} \\
    \hline
    \multirow{2}*{\makecell[l]{\textbf{Ours}}} & \multirow{2}*{\makecell[l]{\textbf{94.39}\\ \; \scalebox{0.9}[0.9]{\textbf{(0.42)}}}} & \multirow{2}*{\makecell[l]{\textbf{63.90}\\ \; \scalebox{0.9}[0.9]{\textbf{(0.24)}}}} & \multirow{2}*{\makecell[l]{\textbf{92.34}\\ \; \scalebox{0.9}[0.9]{\textbf{(0.10)}}}} & \multirow{2}*{\makecell[l]{\textbf{69.37}\\ \; \scalebox{0.9}[0.9]{\textbf{(0.12)}}}} & \multirow{2}*{\makecell[l]{\textbf{98.57}\\ \; \scalebox{0.9}[0.9]{\textbf{(0.28)}}}} & \multirow{2}*{\makecell[l]{\textbf{96.11}\\ \; \scalebox{0.9}[0.9]{\textbf{(0.24)}}}} & \multirow{2}*{\makecell[l]{\textbf{84.60}\\ \; \scalebox{0.9}[0.9]{\textbf{(0.29)}}}} & \multirow{2}*{\makecell[l]{\textbf{63.75}\\ \; \scalebox{0.9}[0.9]{\textbf{(0.32)}}}} & \multirow{2}*{\makecell[l]{\textbf{96.52}\\ \; \scalebox{0.9}[0.9]{\textbf{(0.20)}}}} & \multirow{2}*{\makecell[l]{\textbf{91.23}\\ \; \scalebox{0.9}[0.9]{\textbf{(0.29)}}}} & \multirow{2}*{\makecell[l]{\textit{\textbf{93.28}}\\ \; \scalebox{0.9}[0.9]{\textit{\textbf{(0.26)}}}}} & \multirow{2}*{\makecell[l]{\textit{\textbf{76.87}}\\ \; \scalebox{0.9}[0.9]{\textit{\textbf{(0.24)}}}}} \\
    \multirow{2}*{\makecell[l]{}} & \multirow{2}*{\makecell[l]{}} & \multirow{2}*{\makecell[l]{}} & \multirow{2}*{\makecell[l]{}} & \multirow{2}*{\makecell[l]{}} & \multirow{2}*{\makecell[l]{}} & \multirow{2}*{\makecell[l]{}} & \multirow{2}*{\makecell[l]{}} & \multirow{2}*{\makecell[l]{}} & \multirow{2}*{\makecell[l]{}} & \multirow{2}*{\makecell[l]{}} & \multirow{2}*{\makecell[l]{}} & \multirow{2}*{\makecell[l]{}} \\
    \hline \hline
    \end{tabular}
    \end{center}
\end{table*}

\begin{table*}[!t]
    \begin{center}
    \caption{\fontsize{7.46}{9}\selectfont{Evaluation on Fully Fine-tuned Classification Performance for Different Pre-training Strategies. Each Encoder is Fixed to ViT-B/16. (\%)}}
    \renewcommand{\arraystretch}{1.1}
    \label{table3}
    \begin{tabular}{l c c c c c c c c c c c c}
    \hline \hline
    \multirow{2}*{\makecell[l]{Pre-training \\ Strategy}} & \multicolumn{2}{c}{RFMiD} & \multicolumn{2}{c}{ODIR} &
    \multicolumn{2}{c}{REFUGE} & \multicolumn{2}{c}{MESSIDOR} & \multicolumn{2}{c}{FIVES} & \multicolumn{2}{c}{\textit{Avg.}}\\
    \cline{2-13}
    \multirow{2}*{} & AUC & mAP & AUC & mAP & AUC & mAP & AUC & mAP & AUC & mAP & AUC & mAP \\
    \hline
    \multirow{2}*{\makecell[l]{MAE \cite{he2022masked}}} & \multirow{2}*{\makecell[l]{86.52\\ \; \scalebox{0.9}[0.9]{(0.22)}}} & \multirow{2}*{\makecell[l]{41.50\\ \; \scalebox{0.9}[0.9]{(0.20)}}} & \multirow{2}*{\makecell[l]{83.47\\ \; \scalebox{0.9}[0.9]{(0.48)}}} & \multirow{2}*{\makecell[l]{55.57\\ \; \scalebox{0.9}[0.9]{(0.53)}}} & \multirow{2}*{\makecell[l]{89.16\\ \; \scalebox{0.9}[0.9]{(0.49)}}} & \multirow{2}*{\makecell[l]{80.31\\ \; \scalebox{0.9}[0.9]{(0.56)}}} & \multirow{2}*{\makecell[l]{71.85\\ \; \scalebox{0.9}[0.9]{(1.12)}}} & \multirow{2}*{\makecell[l]{49.48\\ \; \scalebox{0.9}[0.9]{(1.40)}}} & \multirow{2}*{\makecell[l]{95.44\\ \; \scalebox{0.9}[0.9]{(0.22)}}} & \multirow{2}*{\makecell[l]{88.10\\ \; \scalebox{0.9}[0.9]{(0.20)}}} & \multirow{2}*{\makecell[l]{\textit{85.29}\\ \; \scalebox{0.9}[0.9]{\textit{(0.51)}}}} & \multirow{2}*{\makecell[l]{\textit{62.99}\\ \; \scalebox{0.9}[0.9]{\textit{(0.50)}}}} \\
    \multirow{2}*{\makecell[l]{}} & \multirow{2}*{\makecell[l]{}} & \multirow{2}*{\makecell[l]{}} & \multirow{2}*{\makecell[l]{}} & \multirow{2}*{\makecell[l]{}} & \multirow{2}*{\makecell[l]{}} & \multirow{2}*{\makecell[l]{}} & \multirow{2}*{\makecell[l]{}} & \multirow{2}*{\makecell[l]{}} & \multirow{2}*{\makecell[l]{}} & \multirow{2}*{\makecell[l]{}} & \multirow{2}*{\makecell[l]{}} & \multirow{2}*{\makecell[l]{}} \\
    \multirow{2}*{\makecell[l]{CLIP \cite{radford2021learning}}} & \multirow{2}*{\makecell[l]{93.79\\ \; \scalebox{0.9}[0.9]{(0.22)}}} & \multirow{2}*{\makecell[l]{62.26\\ \; \scalebox{0.9}[0.9]{(0.36)}}} & \multirow{2}*{\makecell[l]{92.73\\ \; \scalebox{0.9}[0.9]{(0.38)}}} & \multirow{2}*{\makecell[l]{71.39\\ \; \scalebox{0.9}[0.9]{(0.70)}}} & \multirow{2}*{\makecell[l]{96.75\\ \; \scalebox{0.9}[0.9]{(0.26)}}} & \multirow{2}*{\makecell[l]{94.88\\ \; \scalebox{0.9}[0.9]{(0.38)}}} & \multirow{2}*{\makecell[l]{84.81\\ \; \scalebox{0.9}[0.9]{(0.71)}}} & \multirow{2}*{\makecell[l]{63.19\\ \; \scalebox{0.9}[0.9]{(0.52)}}} & \multirow{2}*{\makecell[l]{97.48\\ \; \scalebox{0.9}[0.9]{(0.35)}}} & \multirow{2}*{\makecell[l]{94.48\\ \; \scalebox{0.9}[0.9]{(0.63)}}} & \multirow{2}*{\makecell[l]{\textit{93.11}\\ \; \scalebox{0.9}[0.9]{\textit{(0.38)}}}} & \multirow{2}*{\makecell[l]{\textit{77.24}\\ \; \scalebox{0.9}[0.9]{\textit{(0.52)}}}} \\
    \multirow{2}*{\makecell[l]{}} & \multirow{2}*{\makecell[l]{}} & \multirow{2}*{\makecell[l]{}} & \multirow{2}*{\makecell[l]{}} & \multirow{2}*{\makecell[l]{}} & \multirow{2}*{\makecell[l]{}} & \multirow{2}*{\makecell[l]{}} & \multirow{2}*{\makecell[l]{}} & \multirow{2}*{\makecell[l]{}} & \multirow{2}*{\makecell[l]{}} & \multirow{2}*{\makecell[l]{}} & \multirow{2}*{\makecell[l]{}} & \multirow{2}*{\makecell[l]{}} \\
    \multirow{2}*{\makecell[l]{DeiT \cite{pmlr-v139-touvron21a}}} & \multirow{2}*{\makecell[l]{93.60\\ \; \scalebox{0.9}[0.9]{(0.18)}}} & \multirow{2}*{\makecell[l]{57.16\\ \; \scalebox{0.9}[0.9]{(0.32)}}} & \multirow{2}*{\makecell[l]{92.25\\ \; \scalebox{0.9}[0.9]{(0.25)}}} & \multirow{2}*{\makecell[l]{69.10\\ \; \scalebox{0.9}[0.9]{(0.33)}}} & \multirow{2}*{\makecell[l]{96.68\\ \; \scalebox{0.9}[0.9]{(0.10)}}} & \multirow{2}*{\makecell[l]{93.51\\ \; \scalebox{0.9}[0.9]{(0.70)}}} & \multirow{2}*{\makecell[l]{84.74\\ \; \scalebox{0.9}[0.9]{(0.28)}}} & \multirow{2}*{\makecell[l]{63.51\\ \; \scalebox{0.9}[0.9]{(0.49)}}} & \multirow{2}*{\makecell[l]{97.54\\ \; \scalebox{0.9}[0.9]{(0.06)}}} & \multirow{2}*{\makecell[l]{94.42\\ \; \scalebox{0.9}[0.9]{(0.28)}}} & \multirow{2}*{\makecell[l]{\textit{92.96}\\ \; \scalebox{0.9}[0.9]{\textit{(0.17)}}}} & \multirow{2}*{\makecell[l]{\textit{75.54}\\ \; \scalebox{0.9}[0.9]{\textit{(0.42)}}}} \\
    \multirow{2}*{\makecell[l]{}} & \multirow{2}*{\makecell[l]{}} & \multirow{2}*{\makecell[l]{}} & \multirow{2}*{\makecell[l]{}} & \multirow{2}*{\makecell[l]{}} & \multirow{2}*{\makecell[l]{}} & \multirow{2}*{\makecell[l]{}} & \multirow{2}*{\makecell[l]{}} & \multirow{2}*{\makecell[l]{}} & \multirow{2}*{\makecell[l]{}} & \multirow{2}*{\makecell[l]{}} & \multirow{2}*{\makecell[l]{}} & \multirow{2}*{\makecell[l]{}} \\
    \multirow{2}*{\makecell[l]{MedCLIP \cite{du2024ret}}} & \multirow{2}*{\makecell[l]{91.47\\ \; \scalebox{0.9}[0.9]{(0.24)}}} & \multirow{2}*{\makecell[l]{54.66\\ \; \scalebox{0.9}[0.9]{(1.00)}}} & \multirow{2}*{\makecell[l]{89.25\\ \; \scalebox{0.9}[0.9]{(0.52)}}} & \multirow{2}*{\makecell[l]{63.15\\ \; \scalebox{0.9}[0.9]{(0.84)}}} & \multirow{2}*{\makecell[l]{96.66\\ \; \scalebox{0.9}[0.9]{(0.47)}}} & \multirow{2}*{\makecell[l]{92.71\\ \; \scalebox{0.9}[0.9]{(0.54)}}} & \multirow{2}*{\makecell[l]{82.38\\ \; \scalebox{0.9}[0.9]{(0.64)}}} & \multirow{2}*{\makecell[l]{60.35\\ \; \scalebox{0.9}[0.9]{(0.70)}}} & \multirow{2}*{\makecell[l]{96.70\\ \; \scalebox{0.9}[0.9]{(0.12)}}} & \multirow{2}*{\makecell[l]{92.13\\ \; \scalebox{0.9}[0.9]{(0.24)}}} & \multirow{2}*{\makecell[l]{\textit{91.29}\\ \; \scalebox{0.9}[0.9]{\textit{(0.40)}}}} & \multirow{2}*{\makecell[l]{\textit{72.60}\\ \; \scalebox{0.9}[0.9]{\textit{(0.66)}}}} \\
    \multirow{2}*{\makecell[l]{}} & \multirow{2}*{\makecell[l]{}} & \multirow{2}*{\makecell[l]{}} & \multirow{2}*{\makecell[l]{}} & \multirow{2}*{\makecell[l]{}} & \multirow{2}*{\makecell[l]{}} & \multirow{2}*{\makecell[l]{}} & \multirow{2}*{\makecell[l]{}} & \multirow{2}*{\makecell[l]{}} & \multirow{2}*{\makecell[l]{}} & \multirow{2}*{\makecell[l]{}} & \multirow{2}*{\makecell[l]{}} & \multirow{2}*{\makecell[l]{}} \\
    \multirow{2}*{\makecell[l]{UniCL \cite{yang2022unified}}} & \multirow{2}*{\makecell[l]{92.47\\ \; \scalebox{0.9}[0.9]{(0.56)}}} & \multirow{2}*{\makecell[l]{55.48\\ \; \scalebox{0.9}[0.9]{(0.39)}}} & \multirow{2}*{\makecell[l]{91.10\\ \; \scalebox{0.9}[0.9]{(0.37)}}} & \multirow{2}*{\makecell[l]{66.34\\ \; \scalebox{0.9}[0.9]{(0.24)}}} & \multirow{2}*{\makecell[l]{95.58\\ \; \scalebox{0.9}[0.9]{(0.21)}}} & \multirow{2}*{\makecell[l]{93.88\\ \; \scalebox{0.9}[0.9]{(0.40)}}} & \multirow{2}*{\makecell[l]{81.74\\ \; \scalebox{0.9}[0.9]{(0.13)}}} & \multirow{2}*{\makecell[l]{60.39\\ \; \scalebox{0.9}[0.9]{(0.08)}}} & \multirow{2}*{\makecell[l]{94.32\\ \; \scalebox{0.9}[0.9]{(0.37)}}} & \multirow{2}*{\makecell[l]{88.26\\ \; \scalebox{0.9}[0.9]{(0.30)}}} & \multirow{2}*{\makecell[l]{\textit{91.04}\\ \; \scalebox{0.9}[0.9]{\textit{(0.33)}}}} & \multirow{2}*{\makecell[l]{\textit{72.87}\\ \; \scalebox{0.9}[0.9]{\textit{(0.28)}}}} \\
    \multirow{2}*{\makecell[l]{}} & \multirow{2}*{\makecell[l]{}} & \multirow{2}*{\makecell[l]{}} & \multirow{2}*{\makecell[l]{}} & \multirow{2}*{\makecell[l]{}} & \multirow{2}*{\makecell[l]{}} & \multirow{2}*{\makecell[l]{}} & \multirow{2}*{\makecell[l]{}} & \multirow{2}*{\makecell[l]{}} & \multirow{2}*{\makecell[l]{}} & \multirow{2}*{\makecell[l]{}} & \multirow{2}*{\makecell[l]{}} & \multirow{2}*{\makecell[l]{}} \\
    \hline
    \multirow{2}*{\makecell[l]{\textbf{Ours}}} & \multirow{2}*{\makecell[l]{\textbf{95.44}\\ \; \scalebox{0.9}[0.9]{\textbf{(0.26)}}}} & \multirow{2}*{\makecell[l]{\textbf{66.47}\\ \; \scalebox{0.9}[0.9]{\textbf{(0.12)}}}} & \multirow{2}*{\makecell[l]{\textbf{93.01}\\ \; \scalebox{0.9}[0.9]{\textbf{(0.11)}}}} & \multirow{2}*{\makecell[l]{\textbf{71.54}\\ \; \scalebox{0.9}[0.9]{\textbf{(0.16)}}}} & \multirow{2}*{\makecell[l]{\textbf{97.91}\\ \; \scalebox{0.9}[0.9]{\textbf{(0.33)}}}} & \multirow{2}*{\makecell[l]{\textbf{95.33}\\ \; \scalebox{0.9}[0.9]{\textbf{(0.25)}}}} & \multirow{2}*{\makecell[l]{\textbf{86.29}\\ \; \scalebox{0.9}[0.9]{\textbf{(0.40)}}}} & \multirow{2}*{\makecell[l]{\textbf{65.02}\\ \; \scalebox{0.9}[0.9]{\textbf{(0.37)}}}} & \multirow{2}*{\makecell[l]{\textbf{98.13}\\ \; \scalebox{0.9}[0.9]{\textbf{(0.11)}}}} & \multirow{2}*{\makecell[l]{\textbf{95.56}\\ \; \scalebox{0.9}[0.9]{\textbf{(0.22)}}}} & \multirow{2}*{\makecell[l]{\textit{\textbf{94.16}}\\ \; \scalebox{0.9}[0.9]{\textit{\textbf{(0.24)}}}}} & \multirow{2}*{\makecell[l]{\textit{\textbf{78.78}}\\ \; \scalebox{0.9}[0.9]{\textit{\textbf{(0.22)}}}}} \\
    \multirow{2}*{\makecell[l]{}} & \multirow{2}*{\makecell[l]{}} & \multirow{2}*{\makecell[l]{}} & \multirow{2}*{\makecell[l]{}} & \multirow{2}*{\makecell[l]{}} & \multirow{2}*{\makecell[l]{}} & \multirow{2}*{\makecell[l]{}} & \multirow{2}*{\makecell[l]{}} & \multirow{2}*{\makecell[l]{}} & \multirow{2}*{\makecell[l]{}} & \multirow{2}*{\makecell[l]{}} & \multirow{2}*{\makecell[l]{}} & \multirow{2}*{\makecell[l]{}} \\
    \hline \hline
    \end{tabular}
    \end{center}
\end{table*}

\subsection{Comparison with Existing Pre-training Strategies}

In this section, we compare the effectiveness of our pre-training strategy with existing pre-training strategies on the same pre-training dataset as our ViLReF. The comparison methods are briefly outlined as follows.
\begin{itemize}
    \item{\textbf{MAE \cite{he2022masked}} enables the model to learn representations via a reconstruction-based pretext task employing mask encoders. Existing experimental results showcase that the model achieves the best performance when the masking ratio is set to $0.75$, meaning it masks $\frac{3}{4}$ of the image areas and reconstructs them.}
    \item{\textbf{CLIP \cite{radford2021learning}} uses InfoNCE Loss to select negative samples from the data and push them apart from positive ones within the feature space.}
    \item{\textbf{DeiT \cite{pmlr-v139-touvron21a}} involves training a convolution-based teacher model and using a distillation token to align the output distribution between the teacher model and the ViT-based student model. This approach helps the student model to learn inductive biases from the teacher model.}
    \item{\textbf{MedCLIP \cite{wang2022medclip}} introduces Semantic Matching Loss to make label similarities indicate the optimization target of feature similarities.}
    \item{\textbf{UniCL \cite{yang2022unified}} incorporates label supervision by considering samples from different modalities that share identical labels as positive pairs, thereby pulling them closer in the feature space.}
\end{itemize}

For all pre-training strategies, we use the same ViT-B/16 vision encoder and the same RoBERTa-wwm-ext-base-chinese text encoder, both initialized from CN-CLIP. Thus, all models started from the same initialization, and the only difference lies in the pre-training strategy. For quantitative evaluation, we employ linear probing, fully fine-tuned classification, and prompt-based out-of-distribution zero-shot classification (hereinafter referred to as prompt-based OOD-ZSC). Linear probing evaluates classification performance by keeping the model parameters fixed and replacing the last layer with a trainable linear layer, examining the quality of knowledge acquired during pre-training. In fully fine-tuned classification, all model parameters are fine-tuned to optimize for downstream classification tasks, which can examine the model's generalization performance. In prompt-based OOD-ZSC, all model parameters are fixed to match multiple images and texts within a mini-batch, testing the transfer performance of the representation learned by the model, and the modality alignment performance between the image and text encoders. To construct the zero-shot text inputs, we translate disease category names into accurate Chinese medical terms. 
The ``others” category in RFMiD and ODIR is excluded due to its lack of specific medical meaning. We input Chinese medical terms into the model directly. For qualitative evaluation, we employ the multi-modal Grad-CAM \cite{selvaraju2017grad} method. This method enables the use of text features as input to backpropagate through the image encoder and generate heatmaps that highlight the regions strongly corresponding to the text.

\subsubsection{Results on linear probing}
We first evaluate the linear probing performance of our pre-training strategy against existing ones across RFMiD, ODIR, REFUGE, MESSIDOR, and FIVES datasets, with results presented in Table \ref{table2}. Our pre-training strategy consistently outperforms the other strategies, enabling the model to achieve the highest AUC and mAP among all test datasets. Notably, for RFMiD, which exhibits a large number of categories and an uneven data distribution, the model pre-trained with our strategy achieves a remarkable AUC of $94.39\%$, which is $1.36$ percentage points (p.p.) higher than the runner-up DeiT-based model. In the DR grading task on MESSIDOR, where distinguishing inter-class features is challenging, our strategy attains the best mAP ($63.75\%$), surpassing the DeiT-based model by $4.08$ p.p.

\begin{table*}[!t]
    \begin{center}
    \caption{\fontsize{7.96}{9}\selectfont{Evaluation on Prompt-based OOD-ZSC Performance for Different Pre-training Strategies. Each Encoder is Fixed to ViT-B/16. (\%)}}
    \renewcommand{\arraystretch}{1.1}
    \label{table4}
    \begin{threeparttable}
    \fontsize{8}{9.5}\selectfont{
    \begin{tabular}{l c c c c c c c c c c c c}
    \hline \hline
    \multirow{2}*{\makecell[l]{Pre-training \\ Strategy}} & \multicolumn{2}{c}{RFMiD} & \multicolumn{2}{c}{ODIR} &
    \multicolumn{2}{c}{REFUGE} & \multicolumn{2}{c}{MESSIDOR} & \multicolumn{2}{c}{FIVES} & \multicolumn{2}{c}{\textit{Avg.}} \\
    \cline{2-13}
    \multirow{2}*{} & AUC & mAP & AUC & mAP & AUC & mAP & AUC & mAP & AUC & mAP & AUC & mAP \\
    \hline
    CLIP & 82.34 & 34.55 & 86.22 & 55.18 & 83.86 & 82.55 & 58.68 & 34.32 & 95.64 & 91.28 & \textit{81.35} & \textit{59.58} \\
    DeiT & 54.24 & 5.64 & 57.25 & 15.65 & 89.70 & 78.93 & 49.81 & 26.20 & 54.31 & 34.72 & \textit{61.06} & \textit{32.23} \\
    MedCLIP & 81.00 & 32.77 & 86.97 & \textbf{59.57} & 86.15 & 85.59 & 54.80 & 28.40 & 91.58 & 85.23 & \textit{80.10} & \textit{58.31} \\
    UniCL & 75.71 & 30.44 & 84.94 & 53.53 & 80.37 & 84.73 & 58.47 & 26.49 & 76.18 & 54.14 & \textit{75.13} & \textit{49.87} \\
    \hline
    \textbf{Ours} & \textbf{87.04} & \textbf{37.68} & \textbf{88.73} & 58.99 & \textbf{93.30} & \textbf{86.79} & \textbf{72.25} & \textbf{47.73} & \textbf{96.07} & \textbf{91.34} & \textit{\textbf{87.48}} & \textit{\textbf{64.51}} \\
    \hline \hline
    \end{tabular}}
    \begin{tablenotes}
    \footnotesize
    \item[*] Obtaining the mean and standard deviation from five repeated runs is not applicable to prompt-based OOD-ZSC because all parameters are fixed.
    \end{tablenotes}
    \end{threeparttable}
    \end{center}
\end{table*}

\begin{figure*}[!t]
    \centerline{\includegraphics[width=0.803\textwidth]{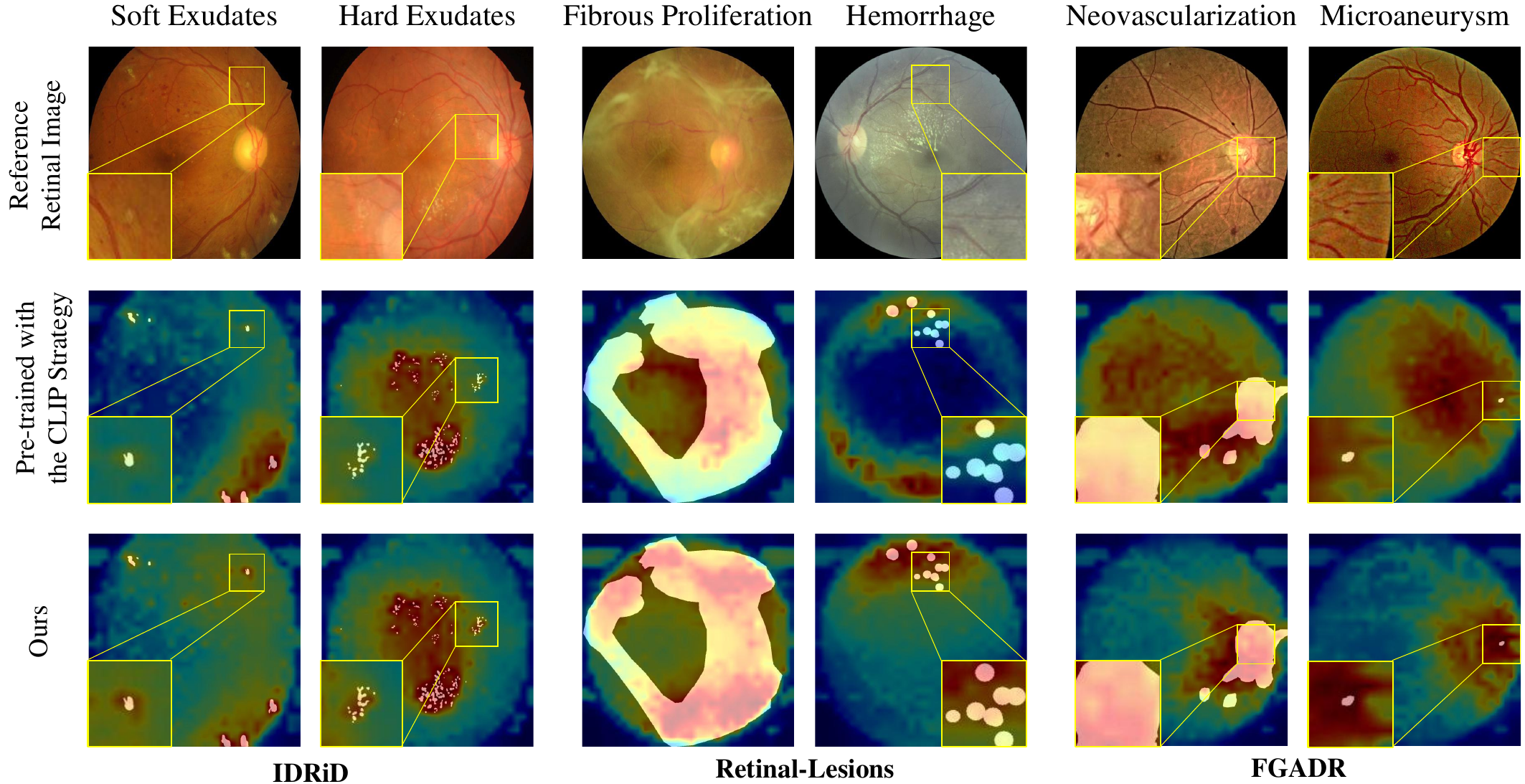}}
    \caption{Visualization of multi-modal activation maps for model pre-trained with the CLIP strategy and ours on IDRiD (exemplar: soft exudates and hard exudates), Retinal-Lesions (exemplar: fibrous proliferation and hemorrhage), and FGADR (exemplar: neovascularization and microaneurysm). The white mask area indicates the ground truth.}
    \label{fig3}
\end{figure*}

The MAE-based model exhibits lower AUC and mAP scores. This can be attributed to the lack of granularity and the high masking ratio of the reconstruction-based pretext task, which hinders the model from learning fine-grained lesion representations effectively. The CLIP strategy may lead to false negative samples, adversely affecting the model performance. However, since many false negatives belong to the ``normal" category, 
the model can still achieve acceptable performance in downstream disease classification tasks. In the DeiT strategy, the presence of false negative samples means the student model optimizes towards a noisy distribution from the teacher model, which does not result in performance improvement. The MedCLIP strategy utilizes Cross-Entropy Loss to align label similarities with feature similarities. While it mitigates the impact of false negatives effectively, a key concern is that label similarity should be regarded as a relative rather than an absolute indicator of feature similarity. For example, two pairs of samples with label similarities $s_{a,b}=1$ and $s_{c,d}=0.5$ can be interpreted as relatively more and less similar, respectively, but this does not imply that their feature distances should follow a strict ratio. Moreover, in medical domain such as ophthalmology, labels often follow a hierarchical structure (e.g., diabetic retinopathy levels from 0 to 4), but the granularity of labeling is typically constrained by the descriptive precision of clinical records (e.g., two samples both labeled as diabetic retinopathy may in fact correspond to moderate and severe stages, respectively). As a result, samples sharing the same label may differ substantially in their pathological manifestations (e.g., local exudates vs. widespread hemorrhage in diabetic retinopathy). Forcing such false positive samples (i.e., samples sharing the same label but differing in semantic content) to be close in the feature space risks blurring clinically meaningful distinctions. For these potential false positive samples in downstream tasks with more subdivided categories, our pre-training strategy eliminates them rather than forcing them to be similar. This approach enables the model to adapt to various downstream tasks effectively. In addition, by treating label similarity as a relative constraint and using it only to adjust the speed of pushing sample pairs further apart within the feature space, our method is inherently less sensitive to label noise, thereby improving robustness to cases with incomplete or missing labels. The UniCL strategy pulls samples with identical labels closer in the feature space. However, such hard label-based modeling lacks precision in complex multi-class scenarios. This coarse granularity can lead to inaccurate alignment.

Our pre-training strategy leverages label information as additional supervision for contrastive learning. By adjusting the speed of pushing sample pairs further apart dynamically within the feature space, it mitigates the impact caused by false negative samples and enables the model to learn more semantically accurate representations.

\subsubsection{Results on fully fine-tuned classification}
\label{fft_vs_lp}
We apply fully fine-tuned classification tasks, comparing our pre-training strategy with existing methods on test datasets. 
The results are illustrated in Table \ref{table3}. From the experimental results, we can draw conclusions consistent with those from the linear probing: the model pre-trained with our strategy demonstrates superior performance across all test datasets. This shows that our pre-training strategy not only enables the model to learn rich and high-quality representations in the pre-training stage but also has strong generalization performance during specific downstream tasks.

\begin{table}[!t]
    \begin{center}
    \caption{Evaluation on Fully Fine-tuned Classification Performance for Ablation Study. Each Encoder is Fixed to Vit-B/16. (\%)}
    \setlength{\tabcolsep}{1.1mm}
    \renewcommand{\arraystretch}{1.1}
    \label{table5}
    \begin{tabular}{l c c c c| c c c c}
    \hline \hline
    & \multicolumn{4}{c|}{AUC} & \multicolumn{4}{c}{mAP}\\
    \cline{2-9}
    \multicolumn{1}{c}{SA} & \ding{55} & \ding{55} & \ding{51} & \ding{52} & \ding{55} & \ding{55} & \ding{51} & \ding{52} \\
    \multicolumn{1}{c}{BE} & \ding{55} & \ding{51} & \ding{55} & \ding{52} & \ding{55} & \ding{51} & \ding{55} & \ding{52} \\
    \hline
    \multirow{2}*{\makecell[l]{RFMiD}} & \multirow{2}*{\makecell[l]{94.32\\  \scalebox{0.9}[0.9]{(0.17)}}} & \multirow{2}*{\makecell[l]{94.46\\  \scalebox{0.9}[0.9]{(0.18)}}} & \multirow{2}*{\makecell[l]{94.82\\  \scalebox{0.9}[0.9]{(0.27)}}} & \multirow{2}*{\makecell[l]{\textbf{95.44}\\  \scalebox{0.9}[0.9]{\textbf{(0.26)}}}} & \multirow{2}*{\makecell[l]{62.74\\  \scalebox{0.9}[0.9]{(0.12)}}} & \multirow{2}*{\makecell[l]{63.33\\  \scalebox{0.9}[0.9]{(0.23)}}} & \multirow{2}*{\makecell[l]{63.22\\  \scalebox{0.9}[0.9]{(0.36)}}} & \multirow{2}*{\makecell[l]{\textbf{66.47}\\  \scalebox{0.9}[0.9]{\textbf{(0.12)}}}}\\
    \multirow{2}*{\makecell[l]{}} & \multirow{2}*{\makecell[l]{}} & \multirow{2}*{\makecell[l]{}} & \multirow{2}*{\makecell[l]{}} & \multirow{2}*{\makecell[l]{}} & \multirow{2}*{\makecell[l]{}} & \multirow{2}*{\makecell[l]{}} & \multirow{2}*{\makecell[l]{}} \\

    \multirow{2}*{\makecell[l]{ODIR}} & \multirow{2}*{\makecell[l]{92.73\\  \scalebox{0.9}[0.9]{(0.38)}}} & \multirow{2}*{\makecell[l]{92.11\\  \scalebox{0.9}[0.9]{(0.34)}}} & \multirow{2}*{\makecell[l]{92.43\\  \scalebox{0.9}[0.9]{(0.13)}}} & \multirow{2}*{\makecell[l]{\textbf{93.01}\\  \scalebox{0.9}[0.9]{\textbf{(0.11)}}}} & \multirow{2}*{\makecell[l]{71.39\\  \scalebox{0.9}[0.9]{(0.70)}}} & \multirow{2}*{\makecell[l]{70.11\\  \scalebox{0.9}[0.9]{(0.55)}}} & \multirow{2}*{\makecell[l]{71.45\\  \scalebox{0.9}[0.9]{(0.56)}}} & \multirow{2}*{\makecell[l]{\textbf{71.54}\\  \scalebox{0.9}[0.9]{\textbf{(0.16)}}}}\\
    \multirow{2}*{\makecell[l]{}} & \multirow{2}*{\makecell[l]{}} & \multirow{2}*{\makecell[l]{}} & \multirow{2}*{\makecell[l]{}} & \multirow{2}*{\makecell[l]{}} & \multirow{2}*{\makecell[l]{}} & \multirow{2}*{\makecell[l]{}} & \multirow{2}*{\makecell[l]{}} \\
    
    \multirow{2}*{\makecell[l]{REFUGE}} & \multirow{2}*{\makecell[l]{96.26\\  \scalebox{0.9}[0.9]{(0.42)}}} & \multirow{2}*{\makecell[l]{96.96\\  \scalebox{0.9}[0.9]{(0.96)}}} & \multirow{2}*{\makecell[l]{97.18\\  \scalebox{0.9}[0.9]{(0.27)}}} & \multirow{2}*{\makecell[l]{\textbf{97.91}\\  \scalebox{0.9}[0.9]{\textbf{(0.33)}}}} & \multirow{2}*{\makecell[l]{94.94\\  \scalebox{0.9}[0.9]{(0.44)}}} & \multirow{2}*{\makecell[l]{94.71\\  \scalebox{0.9}[0.9]{(0.73)}}} & \multirow{2}*{\makecell[l]{95.16\\  \scalebox{0.9}[0.9]{(0.26)}}} & \multirow{2}*{\makecell[l]{\textbf{95.33}\\  \scalebox{0.9}[0.9]{\textbf{(0.25)}}}}\\
    \multirow{2}*{\makecell[l]{}} & \multirow{2}*{\makecell[l]{}} & \multirow{2}*{\makecell[l]{}} & \multirow{2}*{\makecell[l]{}} & \multirow{2}*{\makecell[l]{}} & \multirow{2}*{\makecell[l]{}} & \multirow{2}*{\makecell[l]{}} & \multirow{2}*{\makecell[l]{}} \\

    \multirow{2}*{\makecell[l]{MESSIDOR}} & \multirow{2}*{\makecell[l]{84.81\\  \scalebox{0.9}[0.9]{(0.71)}}} & \multirow{2}*{\makecell[l]{84.52\\  \scalebox{0.9}[0.9]{(0.45)}}} & \multirow{2}*{\makecell[l]{85.37\\  \scalebox{0.9}[0.9]{(0.77)}}} & \multirow{2}*{\makecell[l]{\textbf{86.29}\\  \scalebox{0.9}[0.9]{\textbf{(0.40)}}}} & \multirow{2}*{\makecell[l]{63.19\\  \scalebox{0.9}[0.9]{(0.52)}}} & \multirow{2}*{\makecell[l]{62.09\\  \scalebox{0.9}[0.9]{(0.26)}}} & \multirow{2}*{\makecell[l]{64.12\\  \scalebox{0.9}[0.9]{(0.57)}}} & \multirow{2}*{\makecell[l]{\textbf{65.02}\\  \scalebox{0.9}[0.9]{\textbf{(0.37)}}}}\\
    \multirow{2}*{\makecell[l]{}} & \multirow{2}*{\makecell[l]{}} & \multirow{2}*{\makecell[l]{}} & \multirow{2}*{\makecell[l]{}} & \multirow{2}*{\makecell[l]{}} & \multirow{2}*{\makecell[l]{}} & \multirow{2}*{\makecell[l]{}} & \multirow{2}*{\makecell[l]{}} \\

    \multirow{2}*{\makecell[l]{FIVES}} & \multirow{2}*{\makecell[l]{97.48\\  \scalebox{0.9}[0.9]{(0.35)}}} & \multirow{2}*{\makecell[l]{96.49\\  \scalebox{0.9}[0.9]{(0.17)}}} & \multirow{2}*{\makecell[l]{97.82\\  \scalebox{0.9}[0.9]{(0.20)}}} & \multirow{2}*{\makecell[l]{\textbf{98.13}\\  \scalebox{0.9}[0.9]{\textbf{(0.11)}}}} & \multirow{2}*{\makecell[l]{94.48\\  \scalebox{0.9}[0.9]{(0.63)}}} & \multirow{2}*{\makecell[l]{91.54\\  \scalebox{0.9}[0.9]{(0.40)}}} & \multirow{2}*{\makecell[l]{95.11\\  \scalebox{0.9}[0.9]{(0.33)}}} & \multirow{2}*{\makecell[l]{\textbf{95.56}\\  \scalebox{0.9}[0.9]{\textbf{(0.22)}}}}\\
    \multirow{2}*{\makecell[l]{}} & \multirow{2}*{\makecell[l]{}} & \multirow{2}*{\makecell[l]{}} & \multirow{2}*{\makecell[l]{}} & \multirow{2}*{\makecell[l]{}} & \multirow{2}*{\makecell[l]{}} & \multirow{2}*{\makecell[l]{}} & \multirow{2}*{\makecell[l]{}} \\

    \multirow{2}*{\makecell[l]{\textit{Avg.}}} & \multirow{2}*{\makecell[l]{\textit{93.12}\\  \scalebox{0.9}[0.9]{\textit{(0.41)}}}} & \multirow{2}*{\makecell[l]{\textit{92.91}\\  \scalebox{0.9}[0.9]{\textit{(0.42)}}}} & \multirow{2}*{\makecell[l]{\textit{93.52}\\  \scalebox{0.9}[0.9]{\textit{(0.33)}}}} & \multirow{2}*{\makecell[l]{\textit{\textbf{94.16}}\\  \scalebox{0.9}[0.9]{\textit{\textbf{(0.24)}}}}} & \multirow{2}*{\makecell[l]{\textit{77.35}\\  \scalebox{0.9}[0.9]{\textit{(0.48)}}}} & \multirow{2}*{\makecell[l]{\textit{76.36}\\  \scalebox{0.9}[0.9]{\textit{(0.43)}}}} & \multirow{2}*{\makecell[l]{\textit{77.81}\\  \scalebox{0.9}[0.9]{\textit{(0.42)}}}} & \multirow{2}*{\makecell[l]{\textit{\textbf{78.78}}\\  \scalebox{0.9}[0.9]{\textit{\textbf{(0.22)}}}}}\\
    \multirow{2}*{\makecell[l]{}} & \multirow{2}*{\makecell[l]{}} & \multirow{2}*{\makecell[l]{}} & \multirow{2}*{\makecell[l]{}} & \multirow{2}*{\makecell[l]{}} & \multirow{2}*{\makecell[l]{}} & \multirow{2}*{\makecell[l]{}} & \multirow{2}*{\makecell[l]{}} \\

    \hline \hline
    \end{tabular}
    \end{center}
\end{table}

\begin{table}[!t]
    \begin{center}
    \caption{Evaluation on Prompt-based OOD-ZSC Performance for Ablation Study. Each Encoder is Fixed to Vit-B/16. (\%)}
    \setlength{\tabcolsep}{1.1mm}
    \renewcommand{\arraystretch}{1.1}
    \label{table6}
    \begin{threeparttable}
    \fontsize{8}{9.5}\selectfont{
    \begin{tabular}{l c c c c| c c c c}
    \hline \hline
    & \multicolumn{4}{c|}{AUC} & \multicolumn{4}{c}{mAP}\\
    \cline{2-9}
    \multicolumn{1}{c}{SA} & \ding{55} & \ding{55} & \ding{51} & \ding{52} & \ding{55} & \ding{55} & \ding{51} & \ding{52} \\
    \multicolumn{1}{c}{BE} & \ding{55} & \ding{51} & \ding{55} & \ding{52} & \ding{55} & \ding{51} & \ding{55} & \ding{52} \\
    \hline
    RFMiD & 81.92 & 81.18 & 84.68 & \textbf{87.04} & 35.61 & 36.70 & 37.49 & \textbf{37.68} \\
    ODIR & 86.11 & 81.28 & 88.43 & \textbf{88.73} & 54.26 & 54.99 & 58.36 & \textbf{58.99} \\
    REFUGE & 83.06 & 86.65 & 88.08 & \textbf{93.30} & 81.94 & 80.00 & 84.95 & \textbf{86.79} \\
    MESSIDOR & 58.68 & 65.07 & 71.70 & \textbf{72.25} & 34.32 & 37.09 & 45.74 & \textbf{47.73} \\
    FIVES & 89.98 & 92.17 & 93.70 & \textbf{96.07} & 76.39 & 81.03 & 84.70 & \textbf{91.34} \\
    \textit{Avg.} & \textit{79.95} & \textit{81.27} & \textit{85.32} & \textit{\textbf{87.48}} & \textit{56.50} & \textit{57.96} & \textit{62.24} & \textit{\textbf{64.51}} \\
    \hline \hline
    \end{tabular}}
    \begin{tablenotes}
    \footnotesize
    \item[*] \fontsize{7.96}{9}\selectfont{Obtaining the mean and standard deviation from five repeated runs is not applicable to prompt-based OOD-ZSC because all parameters are fixed.}
    \end{tablenotes}
    \end{threeparttable}
    \end{center}
\end{table}

\begin{figure}[!t]
    \centering
    {\includegraphics[width=\columnwidth]{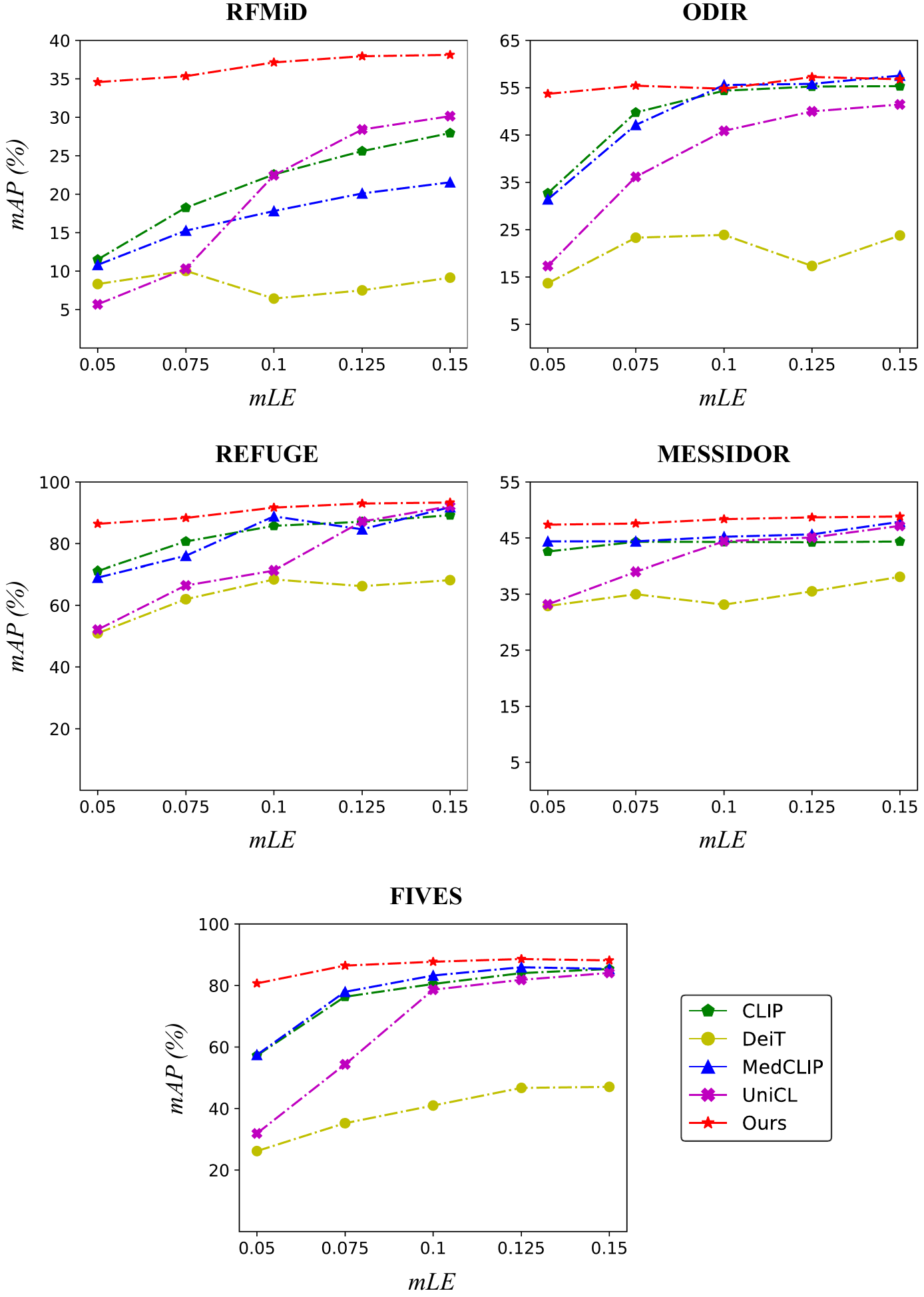}}
    \caption{Comparison of prompt-based OOD-ZSC performance for models trained on datasets with different label entropies.}
    \label{fig4}
\end{figure}

\subsubsection{Results on prompt-based OOD-ZSC}
We apply prompt-based OOD-ZSC to evaluate the feature extraction quality and the visual-language representation alignment performance. The results are presented in Table \ref{table4}. The MAE strategy does not involve text and is therefore excluded from this comparison.

The CLIP strategy introduces noise into the model due to false negatives, complicating the matching of images and texts with the same semantics. The DeiT strategy shows poor performance in cross-modal alignment because the distillation constraint introduces uncertainty. The MedCLIP strategy, which aligns feature similarities with label similarities, enables the model to achieve good performance on some test datasets. For example, it achieves an mAP of $59.57\%$ on ODIR. However, during the pre-training stage, the samples categorized in different grades of DR are assigned to the same label and are optimized to be similar in the feature space, leading to false positive samples in the DR grading task and thus causing lower performance of AUC. Due to the coarse hard label-based modeling, the UniCL strategy achieves lower performance, especially in tasks requiring fine-grained semantic discrimination.

For these potential false positive samples in downstream tasks with more subdivided categories, our pre-training strategy eliminates them rather than forcing them to be similar. This approach enables the model to adapt to various downstream tasks effectively. Our pre-training strategy allows the model to achieve the best overall performance and distinguish diseases in different grades, as demonstrated by the superior AUC of $72.25\%$ and mAP of $47.73\%$ on the DR grading task on MESSIDOR, surpassing the second-best CLIP-based model by $13.57$ p.p. and $13.41$ p.p., respectively.

\begin{figure*}[!t]
    \centerline{\includegraphics[width=\textwidth]{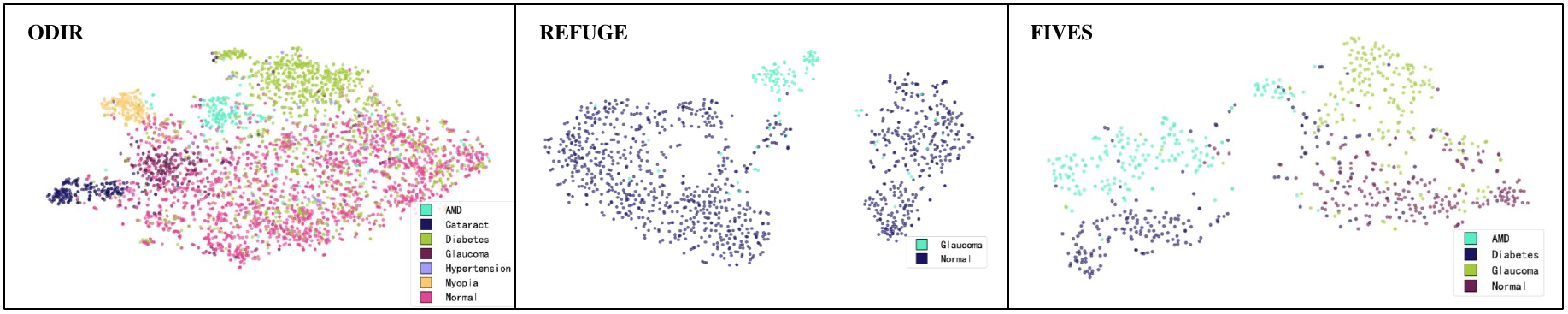}}
    \caption{Visualization of compressed feature distributions encoded by ViLReF on ODIR, REFUGE, and FIVES datasets via t-SNE dimensionality reduction.}
    \label{fig5}
\end{figure*}

\subsubsection{Multi-modal activation map visualization using Grad-CAM}
To demonstrate the interpretability of features extracted by ViLReF more intuitively, we adopt the multi-modal Grad-CAM method to visualize the gradient activation heatmaps. The multi-modal heatmap overlays are drawn on IDRiD, Retinal-Lesions, and FGADR datasets, as shown in Fig. \ref{fig3}, with one sample for each disease. Due to space limitations, we only show the results for ViLReF and the runner-up model pre-trained with the CLIP strategy. Owing to the additional supervision provided by labels and the proposed Weighted Similarity Coupling Loss, ViLReF can learn the lesion features of retinal images more effectively, which can be reflected in the more accurate activation positions of the lesion appearance, location, and extent. Additionally, ViLReF is capable of learning more subtle lesion patterns.

\subsection{Ablation Study}
To evaluate the contribution of each component in our pre-training strategy, we conduct ablation studies using the same pre-training data and fix the image encoder to ViT-B/16 as in previous evaluations. We introduce the speed adjustment of feature similarity (SA) and the batch expansion (BE) to the baseline CN-CLIP. We then compare their performance on fully fine-tuned classification and prompt-based OOD-ZSC.

\subsubsection{Results on fully fine-tuned classification}
The results of fully fine-tuned classification performance are shown in Table \ref{table5}. Without applying SA, only employing BE does not improve the performance significantly, as it does not address the impact of false negative samples. On the other hand, applying SA alone can improve the model's performance significantly. 
When both SA and BE are employed, the model achieves the best performance across all datasets. This indicates that the dynamic memory queues effectively compensate for the vacancies caused by eliminating false negatives. 

\subsubsection{Results on prompt-based OOD-ZSC}
The results of prompt-based OOD-ZSC performance are shown in Table \ref{table6}. From the experimental results, we can draw conclusions consistent with those from the fully fine-tuned task. As expected, the combination of SA and BE leads to the most significant improvement across all datasets. 

These ablation studies highlight the synergistic effect of the two crucial components of our pre-training strategy in improving the model's ability to capture and understand generalizable representations within ophthalmic data, while maintaining strong alignment between visual and language representations across diverse datasets.

\subsection{Further Discussion on Pre-Training Effectiveness}
In this section, we further discuss the effectiveness of our pre-training strategy. We introduce mean label entropy (mLE) to evaluate the impact of training data with different proportions of identical labels on pre-training. Additionally, we visualize the clustering patterns of distributions in the feature space using t-distributed Stochastic Neighbor Embedding (t-SNE) \cite{van2008visualizing}, which helps validate whether our pre-training strategy separates images of different categories correctly and aggregates those of the same category.

\subsubsection{Label Entropy Analysis}
We sample subsets containing 100,000 image-text pairs with different mLEs from the full pre-training dataset to pre-train the model. The mLE is calculated as: 
\begin{multline}
    \label{eq. 14}
    mLE = - \frac{1}{c} \sum_{i} \left[ P(y_i = 1) \log P(y_i = 1) \right. \\ \left. + P(y_i = 0) \log P(y_i = 0) \right].
\end{multline}
When the mLE of a dataset is high, the proportion of identical labels is low; when the mLE is low, the proportion of identical labels is high. The mLE of full pre-training dataset is $0.1423$, while the sampled subsets are $0.05$, $0.075$, $0.1$, $0.125$, and $0.15$, respectively. We investigate the changes in the prompt-based OOD-ZSC performance for models pre-trained with different strategies as the mLE of the pre-training dataset varies. The results are depicted in Fig. \ref{fig4}.

It can be observed that the performance of the five pre-training strategies improves with the increase of mLE. Among them, the CLIP, MedCLIP, and UniCL strategies are affected significantly, indicating that none can mitigate the impact caused by false negative samples effectively. The results of DeiT further demonstrate the aforementioned declaration that the distillation constraint introduces uncertainty. Regardless of the mLE value of the pre-training dataset, our strategy achieves the best prompt-based OOD-ZSC performance. Notably, when the mLE is $0.05$, our strategy outperforms the comparison methods significantly, validating that the additional supervision can guide the optimization direction of contrastive learning effectively, enabling it to learn effective representations from datasets with highly similar labels.

\subsubsection{Feature visualization using t-SNE}
To demonstrate that ViLReF, pre-trained using the proposed strategy, can effectively capture discriminative features in retinal images and possess strong generalization capabilities, we employ the t-SNE method to visualize the multi-class clusters in the ODIR, REFUGE, and FIVES datasets. 
The t-SNE visualization results are shown in Fig. \ref{fig5}. It can be observed that ViLReF effectively clusters retinal images belonging to different diseases. Although retinal image features are highly homogeneous due to the high consistency of image structure and content, ViLReF still achieves good classification results on downstream tasks.

\begin{table*}[!t]
    \begin{center}
    \caption{Evaluation on Fully Fine-tuned Classification Performance for Different Retinal Foundation Models. (\%)}
    \renewcommand{\arraystretch}{1.1}
    \label{table7}
    \begin{tabular}{l l c c c c c c c c c c c c} 
    \hline \hline
    \multirow{2}*{Model} & \multirow{2}*{\makecell[l]{Visual \\ Backbone}} & \multicolumn{2}{c}{RFMiD} & \multicolumn{2}{c}{ODIR} & \multicolumn{2}{c}{REFUGE} & \multicolumn{2}{c}{MESSIDOR} & \multicolumn{2}{c}{FIVES} & \multicolumn{2}{c}{\textit{Avg.}} \\
    \cline{3-14}
    \multirow{2}*{} & \multirow{2}*{} & AUC & mAP & AUC & mAP & AUC & mAP & AUC & mAP & AUC & mAP & AUC & mAP \\
    \hline
    \multirow{2}*{\makecell[l]{FLAIR}} & \multirow{2}*{\makecell[l]{ResNet50}} & \multirow{2}*{\makecell[l]{80.62\\ \; \scalebox{0.9}[0.9]{(0.23)}}} & \multirow{2}*{\makecell[l]{33.56\\ \; \scalebox{0.9}[0.9]{(0.10)}}} & \multirow{2}*{\makecell[l]{83.44\\ \; \scalebox{0.9}[0.9]{(0.46)}}} & \multirow{2}*{\makecell[l]{58.97\\ \; \scalebox{0.9}[0.9]{(0.21)}}} & \multirow{2}*{\makecell[l]{92.23\\ \; \scalebox{0.9}[0.9]{(1.13)}}} & \multirow{2}*{\makecell[l]{89.41\\ \; \scalebox{0.9}[0.9]{(1.46)}}} & \multirow{2}*{\makecell[l]{77.37\\ \; \scalebox{0.9}[0.9]{(1.52)}}} & \multirow{2}*{\makecell[l]{58.00\\ \; \scalebox{0.9}[0.9]{(0.62)}}} & \multirow{2}*{\makecell[l]{91.84\\ \; \scalebox{0.9}[0.9]{(1.54)}}} & \multirow{2}*{\makecell[l]{86.89\\ \; \scalebox{0.9}[0.9]{(1.42)}}} & \multirow{2}*{\makecell[l]{\textit{85.10}\\ \; \scalebox{0.9}[0.9]{\textit{(0.98)}}}} & \multirow{2}*{\makecell[l]{\textit{65.37}\\ \; \scalebox{0.9}[0.9]{\textit{(0.76)}}}} \\
    \multirow{2}*{\makecell[l]{}} & \multirow{2}*{\makecell[l]{}} & \multirow{2}*{\makecell[l]{}} & \multirow{2}*{\makecell[l]{}} & \multirow{2}*{\makecell[l]{}} & \multirow{2}*{\makecell[l]{}} & \multirow{2}*{\makecell[l]{}} & \multirow{2}*{\makecell[l]{}} & \multirow{2}*{\makecell[l]{}} & \multirow{2}*{\makecell[l]{}} & \multirow{2}*{\makecell[l]{}} & \multirow{2}*{\makecell[l]{}} \\

    \multirow{2}*{\makecell[l]{KeepFIT\\ \; \scalebox{0.9}[0.9]{\quad \; (CFP)}}} & \multirow{2}*{\makecell[l]{ResNet50}} & \multirow{2}*{\makecell[l]{86.96\\ \; \scalebox{0.9}[0.9]{(0.42)}}} & \multirow{2}*{\makecell[l]{41.43\\ \; \scalebox{0.9}[0.9]{(0.21)}}} & \multirow{2}*{\makecell[l]{86.74\\ \; \scalebox{0.9}[0.9]{(0.50)}}} & \multirow{2}*{\makecell[l]{59.36\\ \; \scalebox{0.9}[0.9]{(0.19)}}} & \multirow{2}*{\makecell[l]{86.00\\ \; \scalebox{0.9}[0.9]{(1.89)}}} & \multirow{2}*{\makecell[l]{82.82\\ \; \scalebox{0.9}[0.9]{(1.00)}}} & \multirow{2}*{\makecell[l]{80.19\\ \; \scalebox{0.9}[0.9]{(0.29)}}} & \multirow{2}*{\makecell[l]{58.35\\ \; \scalebox{0.9}[0.9]{(0.16)}}} & \multirow{2}*{\makecell[l]{95.00\\ \; \scalebox{0.9}[0.9]{(0.20)}}} & \multirow{2}*{\makecell[l]{88.67\\ \; \scalebox{0.9}[0.9]{(0.45)}}} & \multirow{2}*{\makecell[l]{\textit{86.98}\\ \; \scalebox{0.9}[0.9]{\textit{(0.66)}}}} & \multirow{2}*{\makecell[l]{\textit{66.13}\\ \; \scalebox{0.9}[0.9]{\textit{(0.40)}}}} \\
    \multirow{2}*{\makecell[l]{}} & \multirow{2}*{\makecell[l]{}} & \multirow{2}*{\makecell[l]{}} & \multirow{2}*{\makecell[l]{}} & \multirow{2}*{\makecell[l]{}} & \multirow{2}*{\makecell[l]{}} & \multirow{2}*{\makecell[l]{}} & \multirow{2}*{\makecell[l]{}} & \multirow{2}*{\makecell[l]{}} & \multirow{2}*{\makecell[l]{}} & \multirow{2}*{\makecell[l]{}} & \multirow{2}*{\makecell[l]{}} \\

    \multirow{2}*{\makecell[l]{RETFound\\ \; \scalebox{0.9}[0.9]{\quad \; (CFP)}}} & \multirow{2}*{\makecell[l]{ViT-L/16}} & \multirow{2}*{\makecell[l]{90.43\\ \; \scalebox{0.9}[0.9]{(0.33)}}} & \multirow{2}*{\makecell[l]{53.37\\ \; \scalebox{0.9}[0.9]{(0.15)}}} & \multirow{2}*{\makecell[l]{88.53\\ \; \scalebox{0.9}[0.9]{(0.68)}}} & \multirow{2}*{\makecell[l]{64.51\\ \; \scalebox{0.9}[0.9]{(0.32)}}} & \multirow{2}*{\makecell[l]{94.51\\ \; \scalebox{0.9}[0.9]{(0.98)}}} & \multirow{2}*{\makecell[l]{91.84\\ \; \scalebox{0.9}[0.9]{(1.76)}}} & \multirow{2}*{\makecell[l]{79.03\\ \; \scalebox{0.9}[0.9]{(1.52)}}} & \multirow{2}*{\makecell[l]{56.32\\ \; \scalebox{0.9}[0.9]{(1.34)}}} & \multirow{2}*{\makecell[l]{96.94\\ \; \scalebox{0.9}[0.9]{(0.36)}}} & \multirow{2}*{\makecell[l]{92.87\\ \; \scalebox{0.9}[0.9]{(0.77)}}} & \multirow{2}*{\makecell[l]{\textit{89.89}\\ \; \scalebox{0.9}[0.9]{\textit{(0.77)}}}} & \multirow{2}*{\makecell[l]{\textit{71.78}\\ \; \scalebox{0.9}[0.9]{\textit{(0.87)}}}} \\
    \multirow{2}*{\makecell[l]{}} & \multirow{2}*{\makecell[l]{}} & \multirow{2}*{\makecell[l]{}} & \multirow{2}*{\makecell[l]{}} & \multirow{2}*{\makecell[l]{}} & \multirow{2}*{\makecell[l]{}} & \multirow{2}*{\makecell[l]{}} & \multirow{2}*{\makecell[l]{}} & \multirow{2}*{\makecell[l]{}} & \multirow{2}*{\makecell[l]{}} & \multirow{2}*{\makecell[l]{}} & \multirow{2}*{\makecell[l]{}} \\
    
    \multirow{2}*{\makecell[l]{RET-CLIP}} & \multirow{2}*{\makecell[l]{ViT-B/16}} & \multirow{2}*{\makecell[l]{94.25\\ \; \scalebox{0.9}[0.9]{(0.13)}}} & \multirow{2}*{\makecell[l]{61.21\\ \; \scalebox{0.9}[0.9]{(0.23)}}} & \multirow{2}*{\makecell[l]{92.61\\ \; \scalebox{0.9}[0.9]{(0.32)}}} & \multirow{2}*{\makecell[l]{69.56\\ \; \scalebox{0.9}[0.9]{(0.69)}}} & \multirow{2}*{\makecell[l]{95.82\\ \; \scalebox{0.9}[0.9]{(0.10)}}} & \multirow{2}*{\makecell[l]{94.57\\ \; \scalebox{0.9}[0.9]{(0.43)}}} & \multirow{2}*{\makecell[l]{84.20\\ \; \scalebox{0.9}[0.9]{(0.22)}}} & \multirow{2}*{\makecell[l]{61.77\\ \; \scalebox{0.9}[0.9]{(0.34)}}} & \multirow{2}*{\makecell[l]{97.06\\ \; \scalebox{0.9}[0.9]{(0.26)}}} & \multirow{2}*{\makecell[l]{93.84\\ \; \scalebox{0.9}[0.9]{(0.26)}}} & \multirow{2}*{\makecell[l]{\textit{92.79}\\ \; \scalebox{0.9}[0.9]{\textit{(0.21)}}}} & \multirow{2}*{\makecell[l]{\textit{76.19}\\ \; \scalebox{0.9}[0.9]{\textit{(0.39)}}}} \\
    \multirow{2}*{\makecell[l]{}} & \multirow{2}*{\makecell[l]{}} & \multirow{2}*{\makecell[l]{}} & \multirow{2}*{\makecell[l]{}} & \multirow{2}*{\makecell[l]{}} & \multirow{2}*{\makecell[l]{}} & \multirow{2}*{\makecell[l]{}} & \multirow{2}*{\makecell[l]{}} & \multirow{2}*{\makecell[l]{}} & \multirow{2}*{\makecell[l]{}} & \multirow{2}*{\makecell[l]{}} & \multirow{2}*{\makecell[l]{}} \\

    \hline
    \multirow{2}*{\makecell[l]{\textbf{ViLReF}\\ \; \scalebox{0.9}[0.9]{\quad \; (\textbf{RN50})}}} & \multirow{2}*{\makecell[l]{\textbf{ResNet50}}} & \multirow{2}*{\makecell[l]{94.11\\ \; \scalebox{0.9}[0.9]{(0.27)}}} & \multirow{2}*{\makecell[l]{60.49\\ \; \scalebox{0.9}[0.9]{(0.21)}}} & \multirow{2}*{\makecell[l]{\textbf{93.60}\\ \; \scalebox{0.9}[0.9]{\textbf{(0.08)}}}} & \multirow{2}*{\makecell[l]{\textbf{73.69}\\ \; \scalebox{0.9}[0.9]{\textbf{(0.25)}}}} & \multirow{2}*{\makecell[l]{97.82\\ \; \scalebox{0.9}[0.9]{(0.50)}}} & \multirow{2}*{\makecell[l]{93.83\\ \; \scalebox{0.9}[0.9]{(0.31)}}} & \multirow{2}*{\makecell[l]{\textbf{88.18}\\ \; \scalebox{0.9}[0.9]{\textbf{(0.16)}}}} & \multirow{2}*{\makecell[l]{\textbf{68.89}\\ \; \scalebox{0.9}[0.9]{\textbf{(0.20)}}}} & \multirow{2}*{\makecell[l]{97.55\\ \; \scalebox{0.9}[0.9]{(0.29)}}} & \multirow{2}*{\makecell[l]{93.82\\ \; \scalebox{0.9}[0.9]{(0.71)}}} & \multirow{2}*{\makecell[l]{\textit{\textbf{94.25}}\\ \; \scalebox{0.9}[0.9]{\textit{\textbf{(0.26)}}}}} & \multirow{2}*{\makecell[l]{\textit{78.14}\\ \; \scalebox{0.9}[0.9]{\textit{(0.34)}}}} \\
    \multirow{2}*{\makecell[l]{}} & \multirow{2}*{\makecell[l]{}} & \multirow{2}*{\makecell[l]{}} & \multirow{2}*{\makecell[l]{}} & \multirow{2}*{\makecell[l]{}} & \multirow{2}*{\makecell[l]{}} & \multirow{2}*{\makecell[l]{}} & \multirow{2}*{\makecell[l]{}} & \multirow{2}*{\makecell[l]{}} & \multirow{2}*{\makecell[l]{}} & \multirow{2}*{\makecell[l]{}} & \multirow{2}*{\makecell[l]{}} \\
    
    \multirow{2}*{\makecell[l]{\textbf{ViLReF}\\ \; \scalebox{0.9}[0.9]{\quad \; (\textbf{ViT})}}} & \multirow{2}*{\makecell[l]{\textbf{ViT-B/16}}} & \multirow{2}*{\makecell[l]{\textbf{95.44}\\ \; \scalebox{0.9}[0.9]{\textbf{(0.26)}}}} & \multirow{2}*{\makecell[l]{\textbf{66.47}\\ \; \scalebox{0.9}[0.9]{\textbf{(0.12)}}}} & \multirow{2}*{\makecell[l]{93.01\\ \; \scalebox{0.9}[0.9]{(0.11)}}} & \multirow{2}*{\makecell[l]{71.54\\ \; \scalebox{0.9}[0.9]{(0.16)}}} & \multirow{2}*{\makecell[l]{\textbf{97.91}\\ \; \scalebox{0.9}[0.9]{\textbf{(0.33)}}}} & \multirow{2}*{\makecell[l]{\textbf{95.33}\\ \; \scalebox{0.9}[0.9]{\textbf{(0.25)}}}} & \multirow{2}*{\makecell[l]{86.29\\ \; \scalebox{0.9}[0.9]{(0.40)}}} & \multirow{2}*{\makecell[l]{65.02\\ \; \scalebox{0.9}[0.9]{(0.37)}}} & \multirow{2}*{\makecell[l]{\textbf{98.13}\\ \; \scalebox{0.9}[0.9]{\textbf{(0.11)}}}} & \multirow{2}*{\makecell[l]{\textbf{95.56}\\ \; \scalebox{0.9}[0.9]{\textbf{(0.22)}}}} & \multirow{2}*{\makecell[l]{\textit{94.16}\\ \; \scalebox{0.9}[0.9]{\textit{(0.24)}}}} & \multirow{2}*{\makecell[l]{\textit{\textbf{78.78}}\\ \; \scalebox{0.9}[0.9]{\textit{\textbf{(0.22)}}}}} \\
    \multirow{2}*{\makecell[l]{}} & \multirow{2}*{\makecell[l]{}} & \multirow{2}*{\makecell[l]{}} & \multirow{2}*{\makecell[l]{}} & \multirow{2}*{\makecell[l]{}} & \multirow{2}*{\makecell[l]{}} & \multirow{2}*{\makecell[l]{}} & \multirow{2}*{\makecell[l]{}} & \multirow{2}*{\makecell[l]{}} & \multirow{2}*{\makecell[l]{}} & \multirow{2}*{\makecell[l]{}} & \multirow{2}*{\makecell[l]{}} \\
    \hline \hline
    \end{tabular}
    \end{center}
\end{table*}

\begin{table*}[!t]
    \begin{center}
    \caption{Evaluation on Prompt-based OOD-ZSC Performance for Different Retinal Foundation Models. (\%)}
    \renewcommand{\arraystretch}{1.1}
    \label{table7-2}
    \begin{threeparttable}
    \fontsize{8}{9.5}\selectfont{
    \begin{tabular}{l l c c c c c c c c c c c c}
    \hline \hline
    \multirow{2}*{Model} & \multirow{2}*{\makecell[l]{Visual \\ Backbone}} & \multicolumn{2}{c}{RFMiD} & \multicolumn{2}{c}{ODIR} & \multicolumn{2}{c}{REFUGE} & \multicolumn{2}{c}{MESSIDOR} & \multicolumn{2}{c}{FIVES} & \multicolumn{2}{c}{\textit{Avg.}} \\
    \cline{3-14}
    \multirow{2}*{} & \multirow{2}*{} & AUC & mAP & AUC & mAP & AUC & mAP & AUC & mAP & AUC & mAP & AUC & mAP \\
    \hline
    FLAIR & ResNet50 & 79.01 & 25.40 & 72.24 & 28.03 & 85.65 & 83.87 & 70.23 & 39.67 & 73.31 & 50.94 & \textit{76.09} & \textit{45.58} \\
    KeepFIT (CFP) & ResNet50 & 79.87 & 27.59 & 81.30 & 39.62 & 84.69 & 83.98 & 70.77 & 41.68 & 85.33 & 62.12 & \textit{80.39} & \textit{51.00} \\
    RET-CLIP & ViT-B/16 & 84.91 & 35.16 & 86.92 & 49.34 & 86.51 & 84.94 & 72.44 & 45.75 & 95.38 & 88.70 & \textit{85.23} & \textit{60.78} \\
    \hline
    \textbf{ViLReF (RN50)} & ResNet50 & 83.39 & \textbf{38.07} & 86.93 & 55.56 & 84.84 & 86.32 & 71.02 & \textbf{50.22} & 95.95 & 90.96 & \textit{84.43} & \textit{64.23} \\
    \textbf{ViLReF (ViT)} & ViT-B/16 & \textbf{87.04} & 37.68 & \textbf{88.73} & \textbf{58.99} & \textbf{93.30} & \textbf{86.79} & \textbf{73.25} & 47.73 & \textbf{96.07} & \textbf{91.34} & \textit{\textbf{87.48}} & \textit{\textbf{64.51}} \\
    \hline \hline
    \end{tabular}}
    \begin{tablenotes}
    \footnotesize
    \item[*] \fontsize{7.96}{9}\selectfont{Obtaining the mean and standard deviation from five repeated runs is not applicable to prompt-based OOD-ZSC because all parameters are fixed.}
    \end{tablenotes}
    \end{threeparttable}
    \end{center}
\end{table*}

\subsection{Comparison with SOTA Models}

In this section, we compare the performance of our model with several SOTA foundation models pre-trained on various data domains, including FLAIR \cite{silva2025foundation}, KeepFIT \cite{wu2024mm}, RETFound \cite{zhou2023foundation}, and RET-CLIP \cite{du2024ret}. For fair comparison with existing models, we also train a version of ViLReF using ResNet50 \cite{he2016deep} as the image encoder for the classification downstream tasks. 

\begin{table}[!t]
    \begin{center}
    \caption{Evaluation on Fully Fine-tuned Lesion Segmentation Performance for Different Pre-trained Encoders. Each Encoder is Fixed to ResNet50. (\%)}
    \setlength{\tabcolsep}{1.7mm}
    \renewcommand{\arraystretch}{1.1}
    \label{table8}
    \begin{tabular}{l c c c c c c}
    \hline \hline
    \multirow{3}*{\makecell[l]{Pre-trained \\ Visual Encoder}} & \multicolumn{6}{c}{Hemorrhages}\\
    \cline{2-7}
    \multirow{3}*{} & \multicolumn{2}{c}{IDRiD} & \multicolumn{2}{c}{Retinal-Lesions} & \multicolumn{2}{c}{FGADR}\\
    \cline{2-7}
    \multirow{3}*{} & DSC & IoU & DSC & IoU & DSC & IoU\\
    \hline
    \multirow{2}*{\makecell[l]{CN-CLIP\\ \; \scalebox{0.9}[0.9]{\quad \; (RN50)}}} & \multirow{2}*{\makecell[l]{51.84\\ \; \scalebox{0.9}[0.9]{(0.72)}}} & \multirow{2}*{\makecell[l]{37.64\\ \; \scalebox{0.9}[0.9]{(0.56)}}} & \multirow{2}*{\makecell[l]{34.22\\ \; \scalebox{0.9}[0.9]{(0.04)}}} & \multirow{2}*{\makecell[l]{23.35\\ \; \scalebox{0.9}[0.9]{(0.09)}}} & \multirow{2}*{\makecell[l]{37.77\\ \; \scalebox{0.9}[0.9]{(0.16)}}} & \multirow{2}*{\makecell[l]{26.06\\ \; \scalebox{0.9}[0.9]{(0.11)}}}\\
    \multirow{2}*{\makecell[l]{}} & \multirow{2}*{\makecell[l]{}} & \multirow{2}*{\makecell[l]{}} & \multirow{2}*{\makecell[l]{}} & \multirow{2}*{\makecell[l]{}} & \multirow{2}*{\makecell[l]{}}\\
    \multirow{2}*{\makecell[l]{FLAIR}} & \multirow{2}*{\makecell[l]{48.21\\ \; \scalebox{0.9}[0.9]{(0.89)}}} & \multirow{2}*{\makecell[l]{33.92\\ \; \scalebox{0.9}[0.9]{(0.90)}}} & \multirow{2}*{\makecell[l]{31.76\\ \; \scalebox{0.9}[0.9]{(0.51)}}} & \multirow{2}*{\makecell[l]{21.40\\ \; \scalebox{0.9}[0.9]{(0.30)}}} & \multirow{2}*{\makecell[l]{36.35\\ \; \scalebox{0.9}[0.9]{(0.46)}}} & \multirow{2}*{\makecell[l]{24.85\\ \; \scalebox{0.9}[0.9]{(0.35)}}}\\
    \multirow{2}*{\makecell[l]{}} & \multirow{2}*{\makecell[l]{}} & \multirow{2}*{\makecell[l]{}} & \multirow{2}*{\makecell[l]{}} & \multirow{2}*{\makecell[l]{}} & \multirow{2}*{\makecell[l]{}}\\
    \multirow{2}*{\makecell[l]{KeepFIT\\ \; \scalebox{0.9}[0.9]{\quad \; (CFP)}}} & \multirow{2}*{\makecell[l]{41.29\\ \; \scalebox{0.9}[0.9]{(1.15)}}} & \multirow{2}*{\makecell[l]{29.41\\ \; \scalebox{0.9}[0.9]{(0.71)}}} & \multirow{2}*{\makecell[l]{31.75\\ \; \scalebox{0.9}[0.9]{(0.15)}}} & \multirow{2}*{\makecell[l]{21.39\\ \; \scalebox{0.9}[0.9]{(0.18)}}} & \multirow{2}*{\makecell[l]{36.00\\ \; \scalebox{0.9}[0.9]{(0.22)}}} & \multirow{2}*{\makecell[l]{24.54\\ \; \scalebox{0.9}[0.9]{(0.18)}}}\\
    \multirow{2}*{\makecell[l]{}} & \multirow{2}*{\makecell[l]{}} & \multirow{2}*{\makecell[l]{}} & \multirow{2}*{\makecell[l]{}} & \multirow{2}*{\makecell[l]{}} & \multirow{2}*{\makecell[l]{}}\\
    \hline
    \multirow{2}*{\makecell[l]{\textbf{ViLReF}\\ \; \scalebox{0.9}[0.9]{\quad \; (\textbf{RN50})}}} & \multirow{2}*{\makecell[l]{\textbf{52.65}\\ \; \scalebox{0.9}[0.9]{\textbf{(0.86)}}}} & \multirow{2}*{\makecell[l]{\textbf{38.38}\\ \; \scalebox{0.9}[0.9]{\textbf{(0.60)}}}} & \multirow{2}*{\makecell[l]{\textbf{34.35}\\ \; \scalebox{0.9}[0.9]{\textbf{(0.09)}}}} & \multirow{2}*{\makecell[l]{\textbf{23.44}\\ \; \scalebox{0.9}[0.9]{\textbf{(0.06)}}}} & \multirow{2}*{\makecell[l]{\textbf{38.23}\\ \; \scalebox{0.9}[0.9]{\textbf{(0.48)}}}} & \multirow{2}*{\makecell[l]{\textbf{26.46}\\ \; \scalebox{0.9}[0.9]{\textbf{(0.33)}}}}\\
    \multirow{2}*{\makecell[l]{}} & \multirow{2}*{\makecell[l]{}} & \multirow{2}*{\makecell[l]{}} & \multirow{2}*{\makecell[l]{}} & \multirow{2}*{\makecell[l]{}} & \multirow{2}*{\makecell[l]{}}\\
    \hline \hline
    \multirow{3}*{\makecell[l]{Pre-trained \\ Visual Encoder}} & \multicolumn{6}{c}{Soft Exudates}\\
    \cline{2-7}
    \multirow{3}*{} & \multicolumn{2}{c}{IDRiD} & \multicolumn{2}{c}{Retinal-Lesions} & \multicolumn{2}{c}{FGADR}\\
    \cline{2-7}
    \multirow{3}*{} & DSC & IoU & DSC & IoU & DSC & IoU\\
    \hline
    \multirow{2}*{\makecell[l]{CN-CLIP\\ \; \scalebox{0.9}[0.9]{\quad \; (RN50)}}} & \multirow{2}*{\makecell[l]{59.50\\ \; \scalebox{0.9}[0.9]{(1.13)}}} & \multirow{2}*{\makecell[l]{47.23\\ \; \scalebox{0.9}[0.9]{(1.71)}}} & \multirow{2}*{\makecell[l]{48.22\\ \; \scalebox{0.9}[0.9]{(0.34)}}} & \multirow{2}*{\makecell[l]{35.75\\ \; \scalebox{0.9}[0.9]{(0.24)}}} & \multirow{2}*{\makecell[l]{35.43\\ \; \scalebox{0.9}[0.9]{(0.60)}}} & \multirow{2}*{\makecell[l]{25.75\\ \; \scalebox{0.9}[0.9]{(0.58)}}}\\
    \multirow{2}*{\makecell[l]{}} & \multirow{2}*{\makecell[l]{}} & \multirow{2}*{\makecell[l]{}} & \multirow{2}*{\makecell[l]{}} & \multirow{2}*{\makecell[l]{}} & \multirow{2}*{\makecell[l]{}}\\
    \multirow{2}*{\makecell[l]{FLAIR}} & \multirow{2}*{\makecell[l]{58.02\\ \; \scalebox{0.9}[0.9]{(1.30)}}} & \multirow{2}*{\makecell[l]{45.12\\ \; \scalebox{0.9}[0.9]{(1.77)}}} & \multirow{2}*{\makecell[l]{48.13\\ \; \scalebox{0.9}[0.9]{(0.63)}}} & \multirow{2}*{\makecell[l]{35.58\\ \; \scalebox{0.9}[0.9]{(0.50)}}} & \multirow{2}*{\makecell[l]{33.38\\ \; \scalebox{0.9}[0.9]{(0.98)}}} & \multirow{2}*{\makecell[l]{24.28\\ \; \scalebox{0.9}[0.9]{(0.75)}}} \\
    \multirow{2}*{\makecell[l]{}} & \multirow{2}*{\makecell[l]{}} & \multirow{2}*{\makecell[l]{}} & \multirow{2}*{\makecell[l]{}} & \multirow{2}*{\makecell[l]{}} & \multirow{2}*{\makecell[l]{}}\\
    \multirow{2}*{\makecell[l]{KeepFIT\\ \; \scalebox{0.9}[0.9]{\quad \; (CFP)}}} & \multirow{2}*{\makecell[l]{58.40\\ \; \scalebox{0.9}[0.9]{(0.65)}}} & \multirow{2}*{\makecell[l]{45.69\\ \; \scalebox{0.9}[0.9]{(0.73)}}} & \multirow{2}*{\makecell[l]{47.43\\ \; \scalebox{0.9}[0.9]{(0.46)}}} & \multirow{2}*{\makecell[l]{34.58\\ \; \scalebox{0.9}[0.9]{(0.26)}}} & \multirow{2}*{\makecell[l]{33.47\\ \; \scalebox{0.9}[0.9]{(0.62)}}} & \multirow{2}*{\makecell[l]{24.38\\ \; \scalebox{0.9}[0.9]{(0.61)}}} \\
    \multirow{2}*{\makecell[l]{}} & \multirow{2}*{\makecell[l]{}} & \multirow{2}*{\makecell[l]{}} & \multirow{2}*{\makecell[l]{}} & \multirow{2}*{\makecell[l]{}} & \multirow{2}*{\makecell[l]{}}\\
    \hline
    \multirow{2}*{\makecell[l]{\textbf{ViLReF}\\ \; \scalebox{0.9}[0.9]{\quad \; (\textbf{RN50})}}} & \multirow{2}*{\makecell[l]{\textbf{60.05}\\ \; \scalebox{0.9}[0.9]{\textbf{(1.03)}}}} & \multirow{2}*{\makecell[l]{\textbf{47.86}\\ \; \scalebox{0.9}[0.9]{\textbf{(1.02)}}}} & \multirow{2}*{\makecell[l]{\textbf{49.43}\\ \; \scalebox{0.9}[0.9]{\textbf{(0.51)}}}} & \multirow{2}*{\makecell[l]{\textbf{36.73}\\ \; \scalebox{0.9}[0.9]{\textbf{(0.35)}}}} & \multirow{2}*{\makecell[l]{\textbf{36.11}\\ \; \scalebox{0.9}[0.9]{\textbf{(0.74)}}}} & \multirow{2}*{\makecell[l]{\textbf{26.32}\\ \; \scalebox{0.9}[0.9]{\textbf{(0.63)}}}}\\
    \multirow{2}*{\makecell[l]{}} & \multirow{2}*{\makecell[l]{}} & \multirow{2}*{\makecell[l]{}} & \multirow{2}*{\makecell[l]{}} & \multirow{2}*{\makecell[l]{}} & \multirow{2}*{\makecell[l]{}}\\
    \hline \hline
    \multirow{3}*{\makecell[l]{Pre-trained \\ Visual Encoder}} & \multicolumn{6}{c}{Hard Exudates}\\
    \cline{2-7}
    \multirow{3}*{} & \multicolumn{2}{c}{IDRiD} & \multicolumn{2}{c}{Retinal-Lesions} & \multicolumn{2}{c}{FGADR}\\
    \cline{2-7}
    \multirow{3}*{} & DSC & IoU & DSC & IoU & DSC & IoU\\
    \hline
    \multirow{2}*{\makecell[l]{CN-CLIP\\ \; \scalebox{0.9}[0.9]{\quad \; (RN50)}}} & \multirow{2}*{\makecell[l]{55.18\\ \; \scalebox{0.9}[0.9]{(0.66)}}} & \multirow{2}*{\makecell[l]{40.79\\ \; \scalebox{0.9}[0.9]{(0.41)}}} & \multirow{2}*{\makecell[l]{46.16\\ \; \scalebox{0.9}[0.9]{(0.40)}}} & \multirow{2}*{\makecell[l]{34.35\\ \; \scalebox{0.9}[0.9]{(0.33)}}} & \multirow{2}*{\makecell[l]{44.01\\ \; \scalebox{0.9}[0.9]{(0.17)}}} & \multirow{2}*{\makecell[l]{31.08\\ \; \scalebox{0.9}[0.9]{(0.22)}}}\\
    \multirow{2}*{\makecell[l]{}} & \multirow{2}*{\makecell[l]{}} & \multirow{2}*{\makecell[l]{}} & \multirow{2}*{\makecell[l]{}} & \multirow{2}*{\makecell[l]{}} & \multirow{2}*{\makecell[l]{}}\\
    \multirow{2}*{\makecell[l]{FLAIR}} & \multirow{2}*{\makecell[l]{54.75\\ \; \scalebox{0.9}[0.9]{(0.71)}}} & \multirow{2}*{\makecell[l]{40.90\\ \; \scalebox{0.9}[0.9]{(0.52)}}} & \multirow{2}*{\makecell[l]{47.38\\ \; \scalebox{0.9}[0.9]{(0.30)}}} & \multirow{2}*{\makecell[l]{35.47\\ \; \scalebox{0.9}[0.9]{(0.32)}}} & \multirow{2}*{\makecell[l]{44.30\\ \; \scalebox{0.9}[0.9]{(0.74)}}} & \multirow{2}*{\makecell[l]{31.20\\ \; \scalebox{0.9}[0.9]{(0.47)}}} \\
    \multirow{2}*{\makecell[l]{}} & \multirow{2}*{\makecell[l]{}} & \multirow{2}*{\makecell[l]{}} & \multirow{2}*{\makecell[l]{}} & \multirow{2}*{\makecell[l]{}} & \multirow{2}*{\makecell[l]{}}\\
    \multirow{2}*{\makecell[l]{KeepFIT\\ \; \scalebox{0.9}[0.9]{\quad \; (CFP)}}} & \multirow{2}*{\makecell[l]{54.88\\ \; \scalebox{0.9}[0.9]{(0.38)}}} & \multirow{2}*{\makecell[l]{40.08\\ \; \scalebox{0.9}[0.9]{(0.20)}}} & \multirow{2}*{\makecell[l]{\textbf{48.23}\\ \; \scalebox{0.9}[0.9]{\textbf{(0.23)}}}} & \multirow{2}*{\makecell[l]{\textbf{36.07}\\ \; \scalebox{0.9}[0.9]{\textbf{(0.17)}}}} & \multirow{2}*{\makecell[l]{\textbf{44.93}\\ \; \scalebox{0.9}[0.9]{\textbf{(0.38)}}}} & \multirow{2}*{\makecell[l]{31.77\\ \; \scalebox{0.9}[0.9]{(0.20)}}} \\
    \multirow{2}*{\makecell[l]{}} & \multirow{2}*{\makecell[l]{}} & \multirow{2}*{\makecell[l]{}} & \multirow{2}*{\makecell[l]{}} & \multirow{2}*{\makecell[l]{}} & \multirow{2}*{\makecell[l]{}}\\
    \hline
    \multirow{2}*{\makecell[l]{\textbf{ViLReF}\\ \; \scalebox{0.9}[0.9]{\quad \; (\textbf{RN50})}}} & \multirow{2}*{\makecell[l]{\textbf{56.53}\\ \; \scalebox{0.9}[0.9]{\textbf{(0.18)}}}} & \multirow{2}*{\makecell[l]{\textbf{41.50}\\ \; \scalebox{0.9}[0.9]{\textbf{(0.13)}}}} & \multirow{2}*{\makecell[l]{46.46\\ \; \scalebox{0.9}[0.9]{(0.24)}}} & \multirow{2}*{\makecell[l]{34.59\\ \; \scalebox{0.9}[0.9]{(0.29)}}} & \multirow{2}*{\makecell[l]{44.38\\ \; \scalebox{0.9}[0.9]{(0.22)}}} & \multirow{2}*{\makecell[l]{\textbf{32.00}\\ \; \scalebox{0.9}[0.9]{\textbf{(1.30)}}}}\\
    \multirow{2}*{\makecell[l]{}} & \multirow{2}*{\makecell[l]{}} & \multirow{2}*{\makecell[l]{}} & \multirow{2}*{\makecell[l]{}} & \multirow{2}*{\makecell[l]{}} & \multirow{2}*{\makecell[l]{}}\\
    \hline \hline
    \end{tabular}
    \end{center}
\end{table}

\begin{figure}[!t]
    \centerline{\includegraphics[width=0.736\columnwidth]{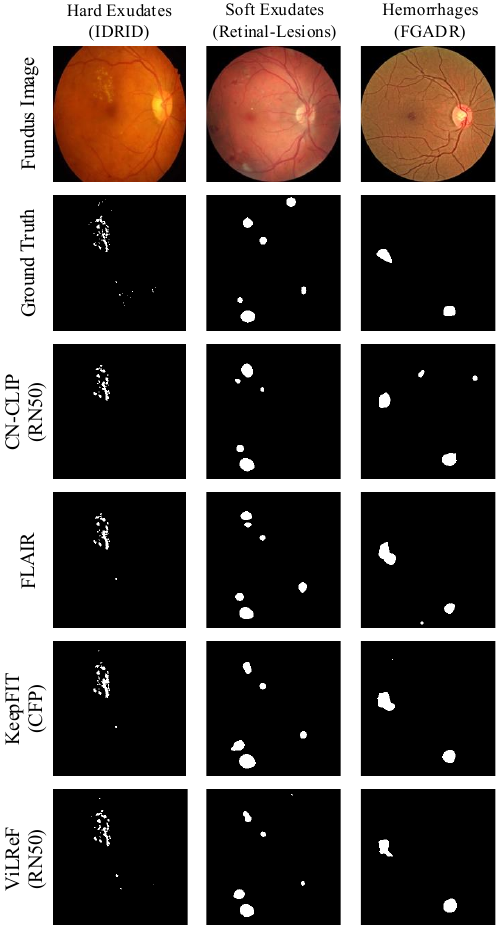}}
    \caption{Quantitative results of fully fine-tuned segmentation for different pre-trained encoders on IDRiD (exemplar: Hard Exudates), Retinal-Lesions (exemplar: Soft Exudates), and FGADR (exemplar: Hemorrhages).}
    \label{fig6}
\end{figure}


\subsubsection{Results on fully fine-tuned classification}
We apply fully fine-tuned classification tasks, comparing ViLReF with SOTA retinal foundation models on test datasets. The results are presented in Table \ref{table7}. 
ViLReF consistently outperforms all SOTA retinal foundation models on fully finetuned classification, confirming its strong generalization ability.

\subsubsection{Results on prompt-based OOD-ZSC}
We further compare prompt-based OOD-ZSC performance of different models to evaluate the feature extraction quality and the visual-language representation alignment performance. For FLAIR and KeepFIT, we follow their template construction protocols and prepend the prompt with “A fundus photograph of” to ensure consistency. For RET-CLIP and our ViLReF, we input Chinese medical terms into the model directly. Note that RETFound is excluded from this comparison as its pre-training does not involve text.

The results are illustrated in Table \ref{table7-2}. ViLReF achieves the best performance in vision-language semantic alignment, particularly the ViT-B/16 version. KeepFIT outperforms FLAIR \cite{silva2025foundation} significantly due to the incorporation of expert knowledge to enhance representation learning. However, both models still show relatively weak performance due to the coarseness of hard label-based modeling. RET-CLIP, which is pre-trained on data of a similar scale to ours, achieves the most comparable classification results to ViLReF. Nevertheless, its use of the CLIP strategy, which introduces noise due to false negatives, results in inferior representation quality and vision-language alignment.

\subsubsection{Results on fully fine-tuned lesion segmentation}
We evaluate the fully fine-tuned lesion segmentation performance on hemorrhages, soft exudates, and hard exudates across IDRiD, Retinal-Lesions, and FGADR datasets. Specifically, we extract intermediate multi-scale features from the visual encoder and connect them to a U-Net-style \cite{ronneberger2015u} decoder via skip connections to construct an end-to-end segmentation framework. During downstream segmentation, both the encoder and the segmentation decoder are fine-tuned jointly. For fair comparison using the same decoder, we use the ResNet50 version of ViLReF and compare its visual encoder with those of CN-CLIP, FLAIR, and KeepFIT, which also use ResNet50. The quantitative results are presented in Table \ref{table8} and the qualitative results are depicted in Fig. \ref{fig6}, covering segmentation performance on three types of lesions: hemorrhages, soft exudates, and hard exudates across IDRiD, Retinal-Lesions, and FGADR datasets. It can be observed that using ViLReF's visual encoder overall enhances the segmentation performance. Both DSC and IoU scores surpass or achieve comparable results with the SOTA retinal foundation models. The results demonstrate that the representations learned by ViLReF are of high quality and have strong generalization ability.

\section{Conclusion}
In this work, we propose ViLReF, a foundation model pre-trained on 451,956 retinal image-diagnostic text report pairs. Expert knowledge is leveraged to guide label extraction, enabling the model to capture subtle but clinically significant visual patterns in retinal images. We present a novel Weighted Similarity Coupling Loss, $\mathcal{L}_{WSC}$, to adjust the speed of pushing sample pairs further apart dynamically within the feature space. Furthermore, a batch expansion module with dynamic memory queues is utilized to mitigate the equivalent batch size reduction incurred by eliminating false negatives. ViLReF achieves superior representation learning performance compared with SOTA foundation models.

Nevertheless, our approach may overlook some fine-grained information from rare categories due to the challenges of label extraction. In future work, we plan to explore advanced approaches for long-tail supervision, including weakly or semi-supervised learning, clustering-based pseudo-labeling, and improved entity recognition for rare classes.

Another limitation is the potential difficulty of applying the expert-knowledge-based report converter to entirely new datasets, as variations in wording, structure, or terminology may reduce its effectiveness. Furthermore, rule-based systems may struggle with nuanced aspects of clinical language, such as severity grading, anatomical location, and negations. Although our dataset was collected from multiple hospitals across China and thus contains diverse textual patterns, improving the adaptability of the converter remains an important future direction. One promising avenue is the development of hybrid approaches (e.g., using the rule-based converter to construct a silver-standard dataset that could then be employed to fine-tune clinical language models), thereby combining expert accuracy with model adaptability. Moreover, we plan to further evaluate ViLReF on more diverse and complex retinal datasets to better assess and enhance its generalizability.

\end{document}